\documentclass[10pt,twocolumn,letterpaper]{article}

\usepackage[pagenumbers]{cvpr} 

\usepackage{graphicx,graphics}
\usepackage{amsmath}
\usepackage{amssymb}
\usepackage{booktabs}

\usepackage{arydshln}
\usepackage{multirow}
\usepackage{multicol}
\usepackage[american]{babel}
\usepackage{microtype}
\usepackage{booktabs,bigstrut}
\usepackage{enumitem}

\graphicspath{{Figure/}}

\usepackage{color, xcolor, colortbl}
\definecolor{citecolor}{HTML}{229954}
\usepackage[pagebackref=true,breaklinks=true,colorlinks,citecolor=citecolor,bookmarks=false]{hyperref}

\newcommand{\bfsection}[1]{\vspace*{0.00cm}\noindent\textbf{#1.}}

\usepackage[capitalize]{cleveref}
\crefname{section}{Sec.}{Secs.}
\Crefname{section}{Section}{Sections}
\Crefname{table}{Table}{Tables}
\crefname{table}{Tab.}{Tabs.}

\def\cvprPaperID{2860} 
\def\confName{CVPR}
\def\confYear{2023}

\begin{document}

    \title{CDDFuse: Correlation-Driven Dual-Branch Feature Decomposition\\for Multi-Modality Image Fusion}

    \author{Zixiang Zhao$^{1,2}$\quad
            Haowen Bai$^{1}$\quad
            Jiangshe Zhang$^{1}$\thanks{Corresponding author.}\quad
            Yulun Zhang$^{2}$\quad \\
            Shuang Xu$^{3,4}$\quad
            Zudi Lin$^{5}$\quad
            Radu Timofte$^{2,6}$\quad
            Luc Van Gool$^{2}$\\[1mm]
		$^{1}$ Xi’an Jiaotong University \quad
            $^{2}$ Computer Vision Lab, ETH Z\"urich\\
            $^{3}$ Research and Development Institute of Northwestern Polytechnical University in Shenzhen\\
            $^{4}$ Northwestern Polytechnical University \quad
            $^{5}$ Harvard University \quad
            $^{6}$ University of W\"urzburg\\
		  {\tt\small zixiangzhao@stu.xjtu.edu.cn, jszhang@mail.xjtu.edu.cn}
}
\maketitle

\begin{abstract}
Multi-modality (MM) image fusion aims to render fused images that maintain the merits of different modalities, \eg, functional highlight and detailed textures. To tackle the challenge in modeling cross-modality features and decomposing desirable modality-specific and modality-shared features, we propose a novel Correlation-Driven feature Decomposition Fusion (\textbf{CDDFuse}) network. Firstly, CDDFuse uses Restormer blocks to extract cross-modality shallow features. We then introduce a dual-branch Transformer-CNN feature extractor with Lite Transformer (LT) blocks leveraging long-range attention to handle low-frequency global features and Invertible Neural Networks (INN) blocks focusing on extracting high-frequency local information. A correlation-driven loss is further proposed to make the low-frequency features correlated while the high-frequency features uncorrelated based on the embedded information. Then, the LT-based global fusion and INN-based local fusion layers output the fused image. Extensive experiments demonstrate that our CDDFuse achieves promising results in multiple fusion tasks, including infrared-visible image fusion and medical image fusion. We also show that CDDFuse can boost the performance in downstream infrared-visible semantic segmentation and object detection in a unified benchmark. The code is available at \url{https://github.com/Zhaozixiang1228/MMIF-CDDFuse}.
\end{abstract}

\begin{figure}[t]
    \centering
    \begin{subfigure}{\linewidth}
        \centering
        \includegraphics[width=\linewidth]{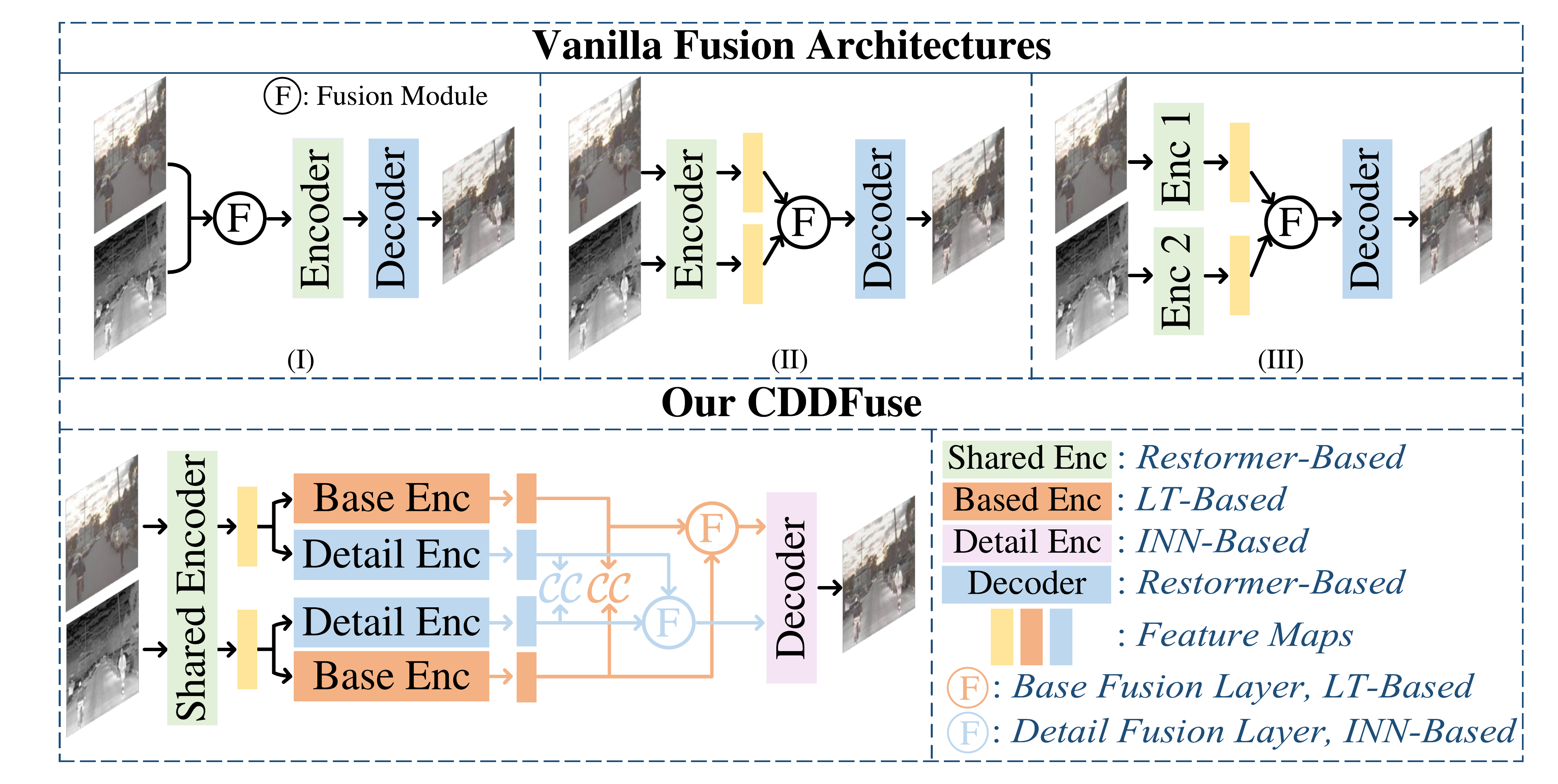}
        \vspace{-1.2em}
        \caption{Existing MMIF methods \vs CDDFuse. The {\em base} and {\em detail} encoders are responsible for extracting global and local features, respectively.}
        \label{fig:introduction1}
    \end{subfigure}
    \begin{subfigure}{\linewidth}
        \centering
        \includegraphics[width=0.9\linewidth]{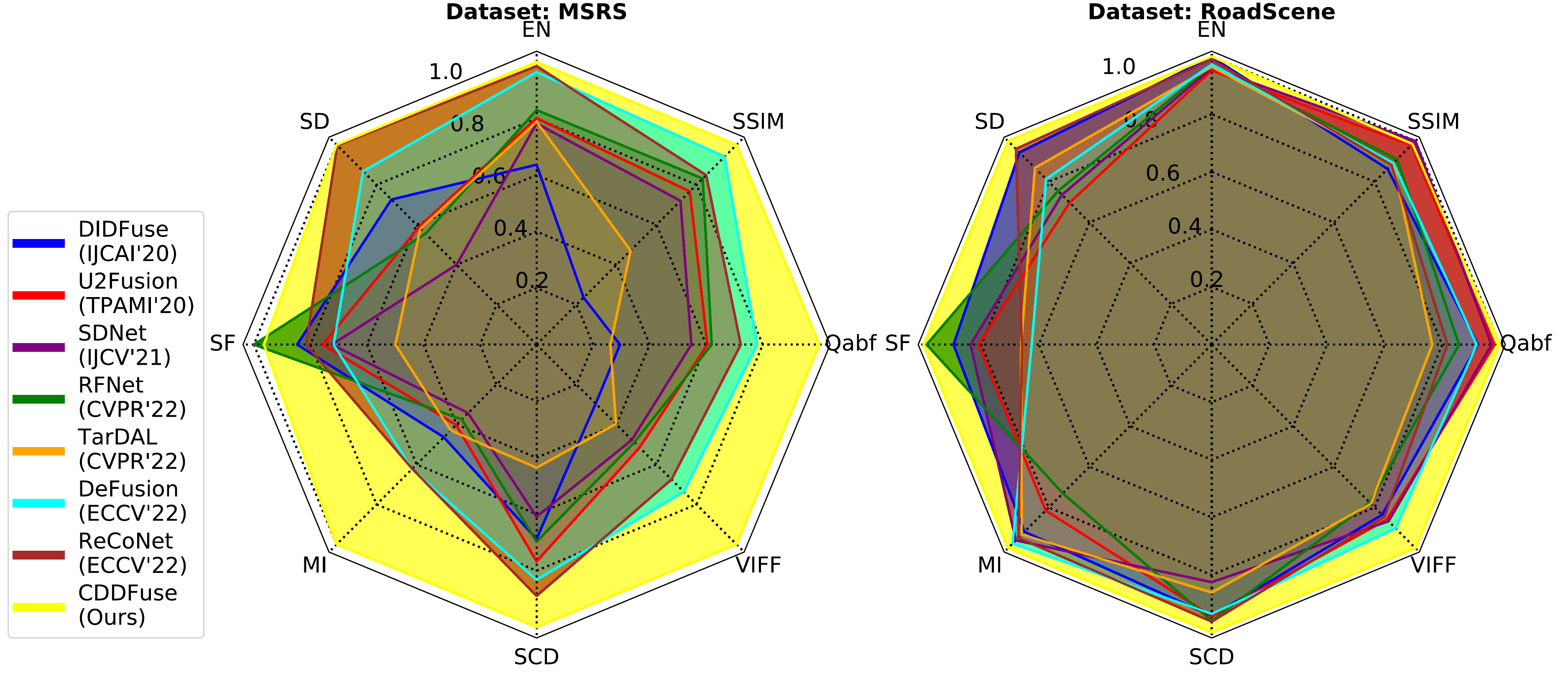}
        \vspace{-0.5em}
        \caption{CDDFuse (highlighted in yellow) achieves the state-of-the-art performance on  MSRS~\cite{DBLP:journals/inffus/TangYZJM22} and RoadScene~\cite{xu2020aaai} in eight metrics.}
        \label{fig:introduction2}
    \end{subfigure}
    \caption{Workflow and performance comparison of our proposed CDDFuse with existing MM image fusion approaches.}
    \label{fig:introduction}
    \vspace{-2em}
\end{figure}

\section{Introduction}\label{sec:1}
Image fusion is a basic image processing topic that aims to generate informative fused images by combining the important information from source ones~\cite{DBLP:journals/sigpro/ZhaoXZLZ20,meher2019a,yao2019spectral,yang2022sir,xu2021stereo}. The fusion targets include digital~\cite{MEF-SSIM2018TCI,zhang2021deep}, multi-modal~\cite{9151265,zhaoijcai2020,DBLP:journals/tcsv/LiuFJLL22} and remote sensing~\cite{DBLP:conf/cvpr/Xu0ZSL021,DBLP:conf/cvpr/BandaraP22,DBLP:conf/icmcs/00010XSHL021} images, \etc. The \textit{Infrared-Visible image Fusion}~(IVF) and \textit{Medical Image Fusion}~(MIF) are two challenging sub-categories of \textit{Multi-Modality Image Fusion} (MMIF), focusing on modeling the cross-modality features from all the sensors and aggregating them into the output images. Specifically, IVF targets fused images that preserve thermal radiation information in the input infrared images and detailed texture information in the input visible images. The fused images can avoid the shortcomings of visible images being sensitive to illumination conditions as well as the infrared images being noisy and low-resolution.
Downstream recognition tasks, \eg, multi-modal saliency detection~\cite{DBLP:conf/cvpr/QinZHGDJ19,Wang2021DualAttention,Liu2019Perceptual}, object detection~\cite{DBLP:journals/corr/abs-2004-10934,liu2021ANP,tang2021robustart} and semantic segmentation~\cite{DBLP:conf/mm/LiuLL021,qin2020forward,qin2022distribution,qin2022bibert} can then benefit from the obtained clearer representations of scenes and objects in IVF images. Similarly, MIF aims to clearly exhibit the abnormalities by fusing multiple medical imaging modalities to reveal comprehensive information to assist diagnosis and treatment~\cite{DBLP:journals/inffus/JamesD14}.

Many methods have been developed to tackle the MMIF challenges in recent years~\cite{DBLP:conf/cvpr/LiuFHWLZL22,DBLP:journals/ijcv/ZhangM21,DBLP:journals/ieeejas/TangDMHM22,DBLP:journals/ieeejas/MaTFHMM22,DBLP:journals/ijcv/MaZJZG19,xu2021classification,xu2021drf}. A common pipeline that demonstrated promising results utilizes CNN-based feature extraction and reconstruction in an Auto-Encoder~(AE) manner~\cite{li2018densefuse,DBLP:journals/inffus/LiWK21,Liang2022ECCV,zhaoijcai2020}. The workflow is illustrated in \cref{fig:introduction1}.
However, existing methods have three main shortcomings.
First, the internal working mechanism of CNNs is difficult to control and interpret, causing insufficient extraction of cross-modality features. For example, in \cref{fig:introduction1}, shared encoders in (\uppercase\expandafter{\romannumeral1}) and (\uppercase\expandafter{\romannumeral2}) cannot distinguish modality-specific features, while the private encoders in (\uppercase\expandafter{\romannumeral3}) ignore features shared by modalities.
Second, the context-independent CNN only extracts local information in a relatively small receptive field, which can hardly extract global information for generating high-quality fused images~\cite{DBLP:conf/iccvw/LiangCSZGT21}. Thus, it is still unclear whether the inductive biases of CNN are capable enough to extract features for all modalities.
Third, the forward propagation of fusion networks often causes the loss of high-frequency information~\cite{ma2019fusiongan,DBLP:journals/tcsv/ZhaoXZLZL22}.
Our work explores a more reasonable paradigm to tackle the challenges in feature extraction and fusion.

First, we aim to add correlation restrictions to the extracted features and limit the solution space, which improves the controllability and interpretability of feature extraction. Our assumption is that, in the MMIF task, the input features of the two modalities are correlated at low frequencies, representing the modality-shared information, while the high-frequency feature is irrelevant and represents the unique characteristics of the respective modalities. Taking IVF as an example, since infrared and visible images come from the same scene, the low-frequency information of the two modalities contains statistical co-occurrences, such as background and large-scale environmental features. On the contrary, the high-frequency information of the two modalities is independent, \eg, the texture and detail information in the visible image and the thermal radiation information in the infrared image. Therefore, we aim to facilitate the extraction of modality-specific and modality-shared features by increasing and decreasing the correlation between low-frequency and high-frequency features, respectively.

Second, from the architectural perspective, Vision Transformers~\cite{DBLP:conf/iclr/DosovitskiyB0WZ21,DBLP:conf/iccv/LiuL00W0LG21,DBLP:conf/cvpr/ZamirA0HK022} recently shows impressive results in computer vision, with self-attention mechanism and global feature extraction. However, Transformer-based methods are computationally expensive, which leaves room for further improvement with considering the efficiency-performance tradeoff of image fusion architectures. Therefore, we propose integrating the advantages of local context extraction and computational efficiency in CNN and the advantages of global attention and long-range dependency modeling in Transformer to complete the MMIF task.

Third, to solve the challenge of losing wanted high-frequency input information, we adopt the building block of \textit{Invertible Neural networks} (INN)~\cite{DBLP:conf/iclr/DinhSB17}. INN was proposed with invertibility by design, which prevents information loss through the mutual generation of input and output features and aligns with our goal of preserving high-frequency features in the fused images.

To this end, we proposed the \textit{Correlation-Driven feature Decomposition Fusion} ({\bf CDDFuse}) model, where modality-specific and modality-shared feature extractions are realized by a dual-branch encoder, with the fused image reconstructed by the decoder. The workflow is shown in \cref{fig:introduction1,fig:Workflow}. Our contributions can be summarized in four aspects:
\begin{itemize}[itemsep=0.1cm,topsep=0.1cm,parsep=0pt]
    \item We propose a dual-branch Transformer-CNN framework for extracting and fusing global and local features, which better reflects the distinct modality-specific and modality-shared features.
    \item We refine the CNN and Transformer blocks for a better adaptation to the MMIF task. Specifically, we are the first to utilize the INN blocks for lossless information transmission and the LT blocks for trading-off fusion quality and computational cost.
    \item We propose a correlation-driven decomposition loss function to enforce the modality shared/specific feature decomposition, which makes the cross-modality base features correlated while decorrelates the detailed high-frequency features in different modalities.
    \item Our method achieves leading image fusion performance for both IVF and MIF. We also present a unified measurement benchmark to justify how the IVF fusion images facilitate downstream MM object detection and semantic segmentation tasks.
\end{itemize}
\section{Related Work}\label{sec:2}
This section briefly reviews the representative works of deep learning~(DL)-based multi-modal image fusion (MMIF) approaches, and the Vision Transformer (LT, Restormer) as well as INN modules employed in CDDFuse.

\subsection{DL-based multi-modal image fusion}
In the era of DL, CNN-based models for MMIF can be categorized into four main classes: generative adversarial network~(GAN)-based models~\cite{ma2019fusiongan,ma2020infrared,DBLP:journals/tim/MaZSLX21}, AE-based models~\cite{li2018densefuse,DBLP:conf/mm/LiuLL021,DBLP:journals/inffus/LiWK21,DBLP:journals/ijcv/ZhangM21}, unified models~\cite{xu2020aaai,DBLP:conf/aaai/ZhangXXGM20,9151265,DBLP:journals/inffus/ZhangLSYZZ20,DBLP:journals/tip/JungKJHS20} and algorithm unrolling models~\cite{DBLP:journals/pami/0002D21,DBLP:journals/tip/GaoDXXD22,DBLP:journals/tcsv/ZhaoXZLZL22,DBLP:journals/corr/abs-2005-08448}. In GAN-based models, GAN~\cite{DBLP:conf/nips/GoodfellowPMXWOCB14,mao2017least,mirza2014conditional} are utilized to simultaneously make the fusion images distributionally similar to inputs and perceptually satisfactory. AE-based methods can be regraded as the DL variant of transformation models with replacing the transformers and inverse transformers with encoders and decoders~\cite{zhaoijcai2020}.
Through cross-task learning, the unified models can alleviate the problems of limited training data and missing ground-truth~\cite{DBLP:conf/aaai/ZhangXXGM20}. Algorithm unrolling models build a bridge between traditional optimization and DL methods, and establish model-driven interpretable CNN frameworks~\cite{DBLP:journals/corr/abs-2104-06977}.
Recently, considering the combination of fusion and downstream pattern recognition tasks, Liu \etal \cite{DBLP:conf/cvpr/LiuFHWLZL22} pioneered the exploration of the combination of image fusion and detection. Then the gradient of the loss function for segmentation~\cite{DBLP:journals/inffus/TangYM22} and detection~\cite{DBLP:conf/mm/SunCZH22} is used to guide the generation of fused images. Liang \etal~\cite{Liang2022ECCV} propose a self-supervised learning framework to complete the fusion task without paired images.
Additionally, adding a pre-processing registration module before the fusion module is proved to solve the misregistration of source images effectively~\cite{DBLP:conf/cvpr/Xu0YLL22,huangreconet,DBLP:conf/ijcai/WangLFL22}. Jiang \etal \cite{DBLP:conf/mm/JiangZ0L22} firstly developed a multi-view and multi-modality fusion based stitching method for comprehensive scene perception.

\subsection{Vision transformer and variants}
Transformer, firstly proposed by Vaswani \etal~\cite{DBLP:conf/nips/VaswaniSPUJGKP17} for natural language processing~(NLP) and ViT~\cite{DBLP:conf/iclr/DosovitskiyB0WZ21} for computer vision.
Then numerous transformer-based models have gained satisfied results in classification \cite{DBLP:conf/icml/TouvronCDMSJ21,DBLP:conf/iccv/LiuL00W0LG21}, object detection \cite{DBLP:conf/eccv/CarionMSUKZ20,DBLP:conf/iclr/ZhuSLLWD21}, segmentation \cite{DBLP:conf/iccv/WangX0FSLL0021,DBLP:conf/cvpr/ZhengLZZLWFFXT021} and multi-modal learning \cite{DBLP:conf/mm/ZhaoZL21,DBLP:conf/mm/JuZLZ20}.
For low-level vision tasks, Transformer combining with multi-task learning~\cite{DBLP:conf/cvpr/Chen000DLMX0021}  and Swin Transformer block~\cite{DBLP:conf/iccv/LiuL00W0LG21,DBLP:conf/iccvw/LiangCSZGT21} has achieved advanced results compared to CNN-based methods. Other advanced networks also obtain competitive results in various inverse problems~\cite{DBLP:journals/corr/abs-2106-06847,DBLP:journals/corr/abs-2203-14186,DBLP:journals/corr/abs-2106-03106,DBLP:journals/corr/abs-2112-10175}.

Considering the large computational overhead of spatial self-attention, Wu \etal~\cite{DBLP:conf/iclr/WuLLLH20} proposed a lightweight LT structure for mobile NLP tasks. Through Long-Short Range Attention and the Flattened feed-forward network, the amount of parameters is largely reduced while maintaining the model performance.
Restormer~\cite{DBLP:conf/cvpr/ZamirA0HK022} improves transformer block by the gated-Dconv network and multi-Dconv head attention transposed modules, which facilitate multi-scale local-global representation learning on high-resolution images. We adopt LT and Restormer blocks into our CDDFuse model.

\subsection{Invertible neural networks}
The invertible neural network is an important module of the Normalized Flow model, a popular kind of generative model~\cite{DBLP:conf/iclr/ArdizzoneKRK19}. It was first proposed by NICE~\cite{DBLP:journals/corr/DinhKB14}, and later the additive coupling layer in NICE was replaced by the coupling layers in RealNVP~\cite{DBLP:conf/iclr/DinhSB17}. Subsequently, 1$\times$1 invertible convolution was used in Glow~\cite{DBLP:conf/nips/KingmaD18}, which can generate realistic high-resolution images. INNs are also applied to classification tasks to save memory and improve the features extraction ability of the backbone~\cite{DBLP:conf/icml/BehrmannGCDJ19,DBLP:conf/iclr/JacobsenSO18,DBLP:conf/nips/GomezRUG17}. Because of its lossless information-preserving property, INNs have been effectively integrated into image processing fields such as image coloring \cite{DBLP:journals/corr/abs-1907-02392}, image hiding~\cite{DBLP:conf/iccv/Jing0XWG21}, image rescaling \cite{DBLP:conf/eccv/XiaoZLWHKBLL20} and image/video super-resolution \cite{DBLP:conf/cvpr/ZhouYHYF022,DBLP:conf/aaai/ZhuLZLLX19}.

\subsection{Comparison with existing approaches}
The existing methods most relevant to our model are the AE-based methods. Compared with the conventional AE methods, our CDDFuse model that extracts local and long-range features with different structures is more reasonable and intuitive than a pure CNN framework. In addition, our proposed correlation-based decomposition loss can effectively suppress redundant information and improve the quality of extracted features than traditional loss functions.

\begin{figure*}[t]
    \centering
    \includegraphics[width=\linewidth]{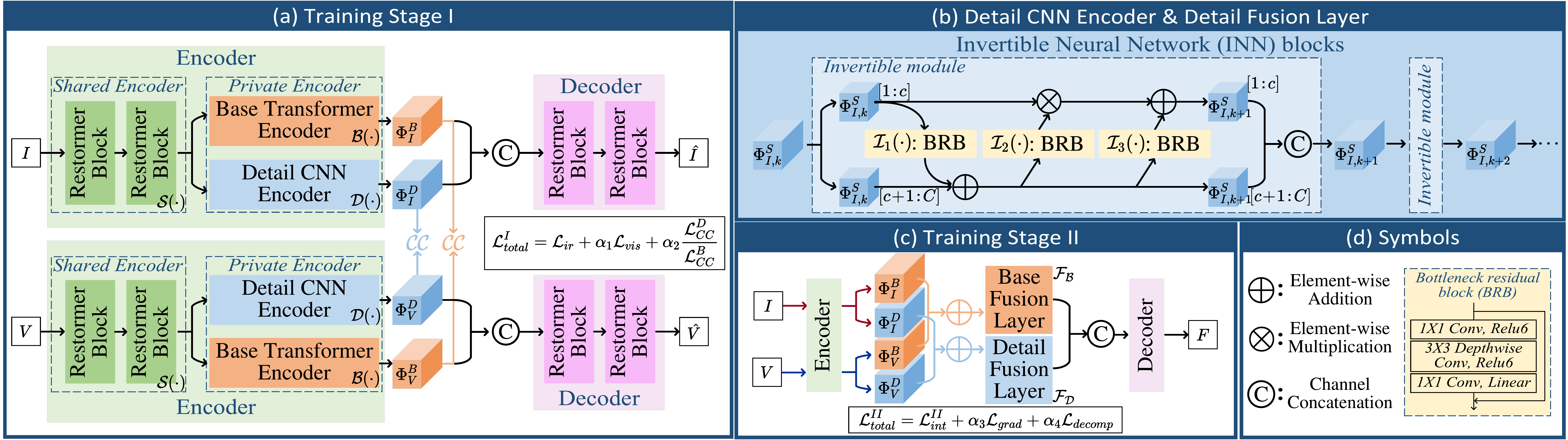}
    \caption{The architecture of our CDDFuse method (IVF as an example).
        \textbf{(a)} The pipeline in training stage \uppercase\expandafter{\romannumeral1}, which aims to train an AE structure for base/detail feature decomposition and reconstructing source images.
        \textbf{(b)} The INN block-based DCE and detail fusion layer, and the BRB block in the affine coupling layers of INN.
        \textbf{(c)} The pipeline in training stage \uppercase\expandafter{\romannumeral2}, which aims to obtain fusion images.
    }
    \label{fig:Workflow}
\end{figure*}

\section{Method}\label{sec:3}
In this section, we first introduce the workflow of CDDFuse and the detailed structure of each module. For simplicity, we denote low-frequency long-range features as the {\em base} features and high-frequency local features as the {\em detail} features in the following discussion.

\subsection{Overview}
Our CDDFuse contains four modules, \ie, a dual-branch encoder for feature extraction and decomposition, a decoder for reconstructing original images (in training stage \uppercase\expandafter{\romannumeral1}) or generating fusion images (in training stage \uppercase\expandafter{\romannumeral2}), and the base/detail fusion layer to fuse the different frequency features, respectively.
The detailed workflow is illustrated in Fig~\ref{fig:Workflow}. Note that CDDFuse is a generic multi-modal image fusion network, and we only take the IVF task as an example to explain the working of CDDFuse.

\subsection{Encoder}
The encoder has three components: the Restormer block~\cite{DBLP:conf/cvpr/ZamirA0HK022}-based \textit{share feature encoder} (SFE), the Lite Transformer (LT) block~\cite{DBLP:conf/iclr/WuLLLH20}-based \textit{base transformer encoder} (BTE) and the Invertible Neural networks (INN) block~\cite{DBLP:conf/iclr/DinhSB17}-based \textit{detail CNN encoder} (DCE). The BTE and DCE together form the \textit{Long-short Range Encoder}.

First, we define some symbols for clarity in formulation. The input paired infrared and visible images are denoted as $I\!\in\!\mathbb{R}^{H\times W}$ and $V\!\in\!\mathbb{R}^{H\times W\times 3}$. The SFE, BTE and DCE are represented by $\mathcal{S}(\cdot)$,  $\mathcal{B}(\cdot)$ and $\mathcal{D}(\cdot)$, respectively.

\bfsection{Share feature encoder}
SFE aims to extracts shallow features $\{\Phi_{I}^S,\Phi_{V}^S\}$ from infrared and visible inputs $\{I,V\}$, \ie,
\begin{equation}\label{equ:loss_SFE}
    \Phi_{I}^S = \mathcal{S}\left(I\right),\  \Phi_{V}^S = \mathcal{S}\left(V\right).
\end{equation}
The reason we choose Restormer block in SFE is that Restormer can extract global features from high-resolution input images by applying self-attention across feature dimension~\cite{DBLP:journals/corr/abs-2111-09881}. Therefore, it can extract cross-modality shallow features without increasing too much computation.
The architecture of Restormer block we use can be referred in supplementary material or the original paper~\cite{DBLP:journals/corr/abs-2111-09881}.

\bfsection{Base transformer encoder}
The BTE is to extract low-frequency base features from the shared features:
\begin{equation}\label{equ:loss_BCE}
    \Phi_{I}^B = \mathcal{B}\left(\Phi_{I}^S\right), \  \Phi_{V}^B = \mathcal{B}\left(\Phi_{V}^S\right).
\end{equation}
where $\Phi_{I}^B$ and $\Phi_{V}^B$ are the base feature of $I$ and $V$, respectively.
In order to extract long-distance dependency features, we use a Transformer with spatial self-attention. Considering to balance the performance and computational efficiency, we use the LT block~\cite{DBLP:conf/iclr/WuLLLH20} as the basic unit of BTE. Through the structure of Flattened feed-forward network which flattens the bottleneck of Transformer blocks, the LT block shrinks the embedding to reduce the number of parameters while preserving the same performance, meeting our expectation.

\bfsection{Detail CNN encoder}
Contrary to BTE, the DCE extracts high-frequency detail information from the shared features, which is formulated as:
\begin{equation}\label{equ:loss_DCE}
    \Phi_{I}^D = \mathcal{D}\left(\Phi_{I}^S\right), \  \Phi_{V}^D = \mathcal{D}\left(\Phi_{V}^S\right).
\end{equation}
Considering that edge and texture information in detail features are very important for image fusion tasks, we hope that the CNN architecture in DCE can preserve as much detail information as possible.
The INN~\cite{DBLP:conf/iclr/DinhSB17} module enables the input information to be better preserved by making its input and output features mutually generated. Thus, it can be regarded as a lossless feature extraction module and is very suitable for use here.
Therefore, we adopt the INN block with affine coupling layers~\cite{DBLP:conf/iclr/DinhSB17,DBLP:journals/tgrs/ZhouFH0LW22}. In each invertible layer, the transformation is:
\begin{equation}\label{equ:INN}
    \begin{aligned}
        \Phi_{I,k+1}^{S}\left[c+1\!:\!C\right] & =\Phi_{I,k}^{S}\left[c+1\!:\!C\right]+\mathcal{I}_1\left(\Phi_{I,k}^{S}\left[1\!:\!c\right]\right),                          \\
        \Phi_{I,k+1}^{S}\left[1\!:\!c\right]   & =\Phi_{I,k}^{S}\left[1\!:\!c\right] \odot \exp \left(\mathcal{I}_2\left(\Phi_{I,k+1}^{S}\left[c+1\!:\!C\right]\right)\right) \\
        & +\mathcal{I}_3\left(\Phi_{I,k+1}^{S}\left[c+1\!:\!C\right]\right),                                                           \\
        \Phi_{I,k+1}^{S}                       & =\mathcal{CAT}\left\{\Phi_{I,k+1}^{S}\left[1\!:\!c\right],\Phi_{I,k+1}^{S}\left[c+1\!:\!C\right]\right\}
    \end{aligned}
\end{equation}
where the $\odot$ is the Hadamard product, $\Phi_{I,k}^{S}\left[1\!:\!c\right]\!\in\!\mathbb{R}^{h\times w\times c}$ is the 1st to the $c$th channels of input feature for the $k$th invertible layer ($k=1,\cdots,K$), $\mathcal{CAT}(\cdot)$ is the channel concatenation operation and $\mathcal{I}_i$ ($i=1,\cdots,3$) are the arbitrary mapping functions. The calculation details can be seen in Fig~\ref{fig:Workflow}(d) and the supplementary meterial. In each invertible layer, $\mathcal{I}_i$ can be set to any mapping without affecting the lossless information transmission in this invertible layer. Considering the trade-off between computational consumption and feature extraction ability, we employ bottleneck residual block (BRB) block in MobileNetV2~\cite{DBLP:conf/cvpr/SandlerHZZC18} as $\mathcal{I}_i$. Finally, $\Phi_{I}^{D} = \Phi_{I,K}^{S}$ and $\Phi_{V}^{D}$ can be obtained in the same way, by replacing the subscript in \cref{equ:INN} from $I$ to $V$.
\subsection{Fusion layer}
The function of the base/detail fusion layer is to fuse base/detail features, respectively. Considering the inductive bias for base/detail feature fusion should be similar to base/detail feature extraction in the encoder, we employ LT and INN blocks for the base and detail fusion layer, where:
\begin{equation}\label{equ:loss_FL}
    \Phi^B = \mathcal{F_B}\left(\Phi_{I}^B,\Phi_{I}^B\right),\ \Phi^D = \mathcal{F_D}\left(\Phi_{I}^D,\Phi_{V}^D\right),
\end{equation}
$\mathcal{F_B}$ and $\mathcal{F_D}$ are the base and detail fusion layer, respectively.
\subsection{Decoder}
In the decoder $\mathcal{DC}(\cdot)$, the decomposed features are concatenated in the channel dimension as the input, and the original image (training stage \uppercase\expandafter{\romannumeral1}) or the fused image (training stage \uppercase\expandafter{\romannumeral2}) is the output of the decoder, which is formulated as:
\begin{equation}\label{equ:Decoder}
    \begin{aligned}
        & \text{Stage \uppercase\expandafter{\romannumeral1}: } & \hat I=\mathcal{DC}\left(\Phi_{I}^B,\Phi_{I}^D\right), &\, \hat V=\mathcal{DC}\left(\Phi_{V}^B,\Phi_{V}^D\right); \\
        & \text{Stage \uppercase\expandafter{\romannumeral2}: } & F     =\mathcal{DC}\left(\Phi^B,\Phi^D\right).         &
    \end{aligned}
\end{equation}
Since the inputs here involving cross-modality and multi-frequency features, we keep the decoder structure consistent with the design of SFE, \ie, using the Restormer block as the basic unit of the decoder.
\subsection{Two-stage training}
A big challenge of the MMIF task is that due to its lack of ground truth, advanced supervised learning methods are ineffective. Here, inspired by \cite{DBLP:journals/inffus/LiWK21}, we use a two-stage learning scheme to train our CDDFuse end-to-end.

\bfsection{Training stage \uppercase\expandafter{\romannumeral1}}
In the training stage \uppercase\expandafter{\romannumeral1}, the paired infrared and visible images $\{I,V\}$ are input into the SFE to extracts shallow features $\{\Phi_{I}^S, \Phi_{V}^S\}$. Then the LT block-based BTE and the INN-based DCE are employed to extract low-frequency base feature $\{\Phi_{I}^B, \Phi_{V}^B\}$ and high-frequency detail feature $\{\Phi_{I}^D, \Phi_{V}^D\}$ for the two different modalities, respectively. After that, the base and detail features of infrared $\{\Phi_{I}^B,\Phi_{I}^D\}$ (or visible $\{\Phi_{V}^B, \Phi_{V}^D\}$) images are concatenated and input into the decoder to reconstruct the original infrared image $\hat I$ (or visible image $\hat V$).

\bfsection{Training stage \uppercase\expandafter{\romannumeral2}}
In the training stage \uppercase\expandafter{\romannumeral2}, the paired infrared and visible images $\{I,V\}$ are input into a nearly well-trained Encoder to obtain the decomposition features. Then
the decomposed base features $\{\Phi_{I}^B, \Phi_{V}^B\}$ and detail features $\{\Phi_{I}^D, \Phi_{V}^D\}$ are input into the fusion layer $\mathcal{F_B}$ and $\mathcal{F_D}$, respectively. At last, the fused features $\{\Phi^B, \Phi^D\}$ are input into the decoder to obtain the fused image $F$.

\bfsection{Training losses}
In training stage \uppercase\expandafter{\romannumeral1}, the total loss $\mathcal{L}_{total}^{\uppercase\expandafter{\romannumeral1}}$ is:
\begin{equation}\label{equ:loss1}
    \mathcal{L}_{total}^{\uppercase\expandafter{\romannumeral1}}
    = \mathcal{L}_{ir} + \alpha_1\mathcal{L}_{vis} + \alpha_2\mathcal{L}_{decomp},
\end{equation}
where $\mathcal{L}_{ir}$ and $\mathcal{L}_{vis}$ are the reconstruction losses for infrared and visible images, $\mathcal{L}_{decomp}$ is the feature decomposition loss, and $\alpha_1$ as well as $\alpha_2$ are the tuning papameters.
The reconstruction losses mainly ensure that the information contained in the images is not lost during the encoding and decoding process, \ie
\begin{equation}\label{equ:loss_ir}
    \mathcal{L}_{ir} =\mathcal{L}_{int}^{\uppercase\expandafter{\romannumeral1}}(I,\hat{I}) + \mu \mathcal{L}_{SSIM}(I,\hat{I}),
\end{equation}
where $\mathcal{L}_{int}^{\uppercase\expandafter{\romannumeral1}}\!=\!\| I-\hat{I} \|_2^2$ and $\mathcal{L}_{SSIM}(I,\hat{I})\!=\! 1\!-\!SSIM(I,\hat I)$. $SSIM(\cdot,\cdot)$ is the structural similarity index~\cite{wang2004image}. $\mathcal{L}_{vis}$ can be obtained in the same way.
Additionally, our proposed feature decomposition loss $\mathcal{L}_{decomp}$ is:
\begin{equation}\label{equ:loss_d}
    \mathcal{L}_{decomp} = \frac{\left(\mathcal{L}_{CC}^D\right)^2}{\mathcal{L}_{CC}^B} = \frac{\left(\mathcal{CC}\left(\Phi_{I}^D,\Phi_{V}^D\right)\right)^2}{\mathcal{CC}\left(\Phi_{I}^B,\Phi_{V}^B\right)+\epsilon}
\end{equation}
where $\mathcal{CC}\left(\cdot, \cdot\right)$ is the correlation coefficient operator, and $\epsilon$ here is set to 1.01 to ensure this term always being positive.

The motivation of this loss term is that, according to our MMIF assumption, the decomposed features $\{\Phi_{I}^B, \Phi_{V}^B\}$ will contain more modality-shared information, such as background and large-scale environment, so they are often highly correlated.
In contrast, $\{\Phi_{I}^D, \Phi_{V}^D\}$ represents the texture and detail information in $V$ and the thermal radiation as well as clear edge information in $I$, which is modality-specific. Thus, the feature maps are less correlated.
Empirically, under the guidance of $\mathcal{L}_{decomp}$ in gradient descent, $\mathcal{L}_{CC}^D$ gradually approaches 0 and $\mathcal{L}_{CC}^B$ becomes larger, which satisfies our intuition for feature decomposition.
The visualization of the decomposition effect will be shown in \cref{fig:fvisual}.

Subsequently in training stage \uppercase\expandafter{\romannumeral2}, inspaired by \cite{DBLP:journals/inffus/TangYM22}, the total loss becomes:
\begin{equation}\label{equ:loss2}
    \mathcal{L}_{total}^{\uppercase\expandafter{\romannumeral2}}
    = \mathcal{L}_{int}^{\uppercase\expandafter{\romannumeral2}} + \alpha_3\mathcal{L}_{grad} + \alpha_4\mathcal{L}_{decomp},
\end{equation}
where $\mathcal{L}_{int}^{\uppercase\expandafter{\romannumeral2}}\!=\!\frac{1}{HW}\!\|I_f\!-\!\max (I_{ir}, I_{vis})\|_1$ and $\mathcal{L}_{grad}\!=\!\frac{1}{HW}\!\|\left|\nabla I_f\right|\!-\!\max (\left|\nabla I_{i r}\right|\!,\!\left|\nabla I_{v i s}\right|)\|_1$.
$\nabla$ indicates the Sobel gradient operator. $\alpha_3$ and $\alpha_4$ are the tuning parameters.
\begin{figure*}[t]
    \centering
    \includegraphics[width=0.85\linewidth]{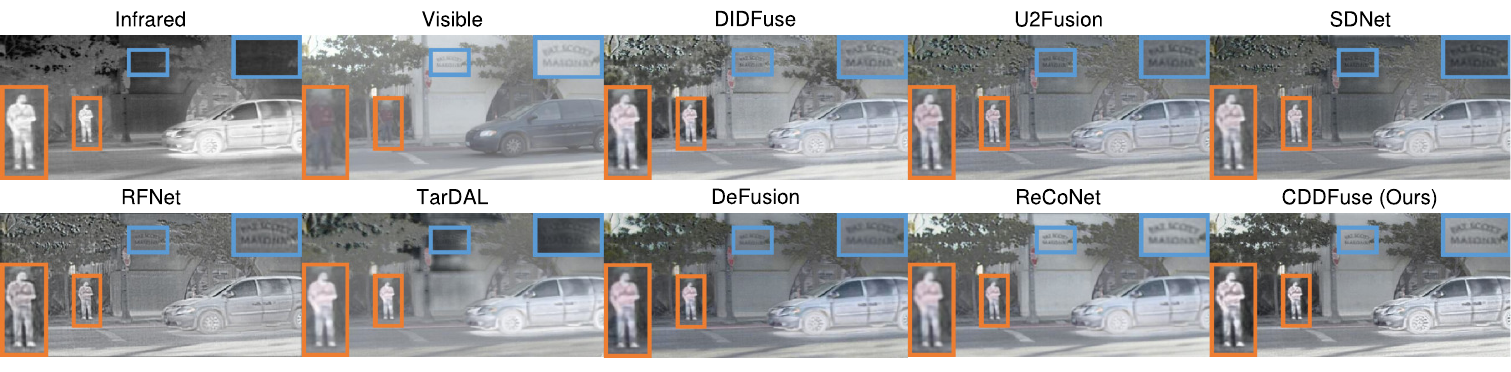}
    \caption{Visual comparison for ``FLIR\_04602'' in RoadScene IVF dataset.}
    \label{fig:IVF1}
\end{figure*}
\begin{figure*}[t]
    \centering
    \vspace{-1em}
    \includegraphics[width=0.85\linewidth]{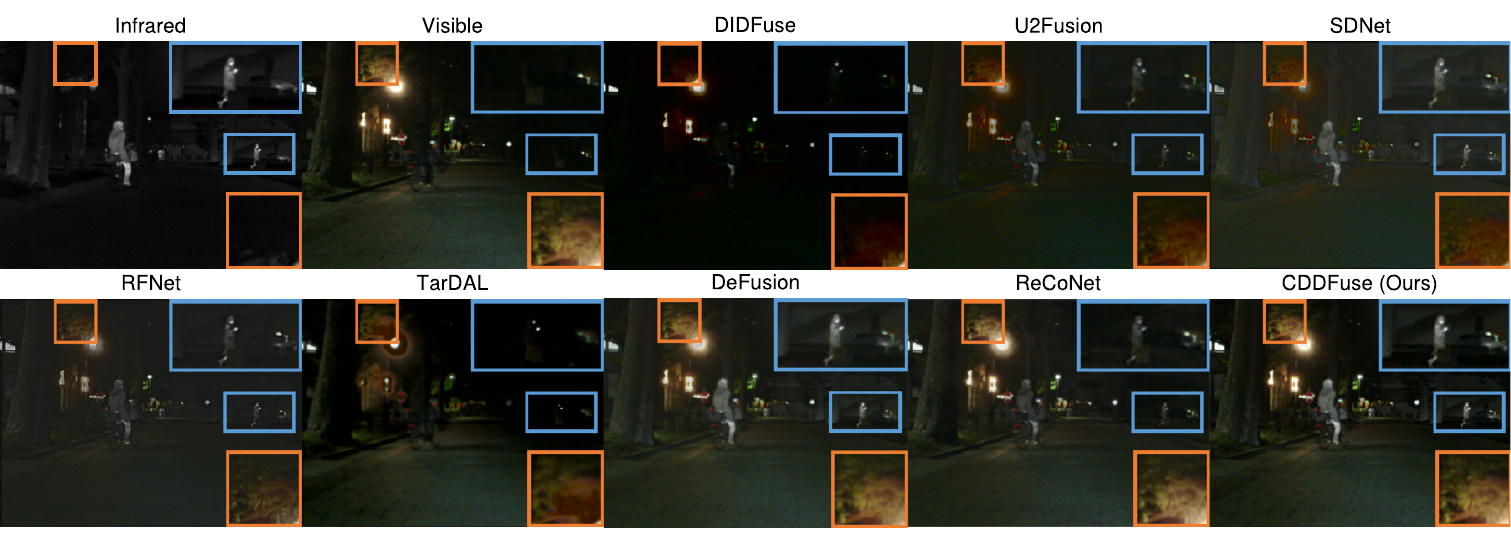}
    \caption{Visual comparison for ``00706N'' in MSRS IVF dataset.}
    \label{fig:IVF2}
    \vspace{-1em}
\end{figure*}

\section{Infrared and visible image fusion}\label{sec:experiment}
Here we elaborate the implementation and configuration details of our networks for the IVF task. Experiments are conducted to show the performance of our models and the rationality of network structures.
\subsection{Setup}
\bfsection{Datasets and metrics}
IVF experiments use three popular benchmarks to verify our fusion model, i.e., MSRS~\cite{DBLP:journals/inffus/TangYZJM22}, RoadScene~\cite{xu2020aaai}, and TNO~\cite{TNO}.
We train our network on MSRS training set (1083 pairs) and 50 pairs in RoadScene are used for validation. MSRS test set (361 pairs), RoadScene (50 pairs) and TNO (25 pairs) are employed as test datasets, which the fusion performance can be verified comprehensively. Note that fine-tuning is not applied to the RoadScene and TNO datasets to verify the generalization performance of the fusion models.

We use eight metrics to quantitatively measure the fusion results: entropy (EN), standard deviation (SD), spatial frequency (SF), mutual information (MI), sum of the correlations of differences (SCD), visual information fidelity (VIF), $Q^{AB/F}$ and structural similarity index measure (SSIM). Higher metrics indicate that a fusion image is better. The details of these metrics can be found in~\cite{ma2019infrared}.

\bfsection{Implement details}
Our experiments are carried out on a machine with two NVIDIA GeForce RTX 3090 GPUs. The training samples are randomly cropped into 128$\times$128 patches in the preprocessing stage. The number of epochs for training is set to 120 with 40 and 80 epochs in the first and second stages, respectively. The batch size is set to 16. We adopt the Adam optimizer with the initial learning rate set to $10^{-4}$ and decreasing by 0.5 every 20 epochs.
For the network hyperparameters setting, the number of Restormer blocks in SFE is 4, with 8 attention heads and 64 dimensions. The dimension of the LT block in BTE is also 64 with 8 attention heads. The configuration of decoder is the same as encoder.
As for loss functions Eq.~(\ref{equ:loss1}) and (\ref{equ:loss2}), $\alpha_1$ to $\alpha_4$ are set to 1, 2, 10, and 2, in order to keep the same order of magnitude for each term.

\subsection{Comparison with SOTA methods}\label{sec:SOTA}
In this section, we test CDDFuse on the three test sets and compare the fusion results with the state-of-the-art methods including
DIDFuse~\cite{zhaoijcai2020},
U2Fusion~\cite{9151265},
SDNet~\cite{DBLP:journals/ijcv/ZhangM21},
RFNet~\cite{DBLP:conf/cvpr/Xu0YLL22},
TarDAL~\cite{DBLP:conf/cvpr/LiuFHWLZL22},
DeFusion~\cite{Liang2022ECCV} and
ReCoNet~\cite{huangreconet}.

\bfsection{Qualitative comparison}
We show the qualitative comparison in \cref{fig:IVF1,fig:IVF2}.
Obviously, our method better integrates thermal radiation information in infrared images and detailed textures in visible images.
Objects in dark regions are clearly highlighted, so that foreground targets can be easily distinguished from the background. Additionally, background details that are difficult to identify due to the low illumination have clear edges and abundant contour information, which help us understand the scene better.

\bfsection{Quantitative comparison}
Afterward, eight metrics are employed to quantitatively compare the above results, which are displayed in \cref{tab:Quantitative}. Our method has excellent performance on almost all metrics, proving that our method is suitable for various kinds of illumination and target categories.

\begin{figure}[t]
    \centering
    \includegraphics[width=\linewidth]{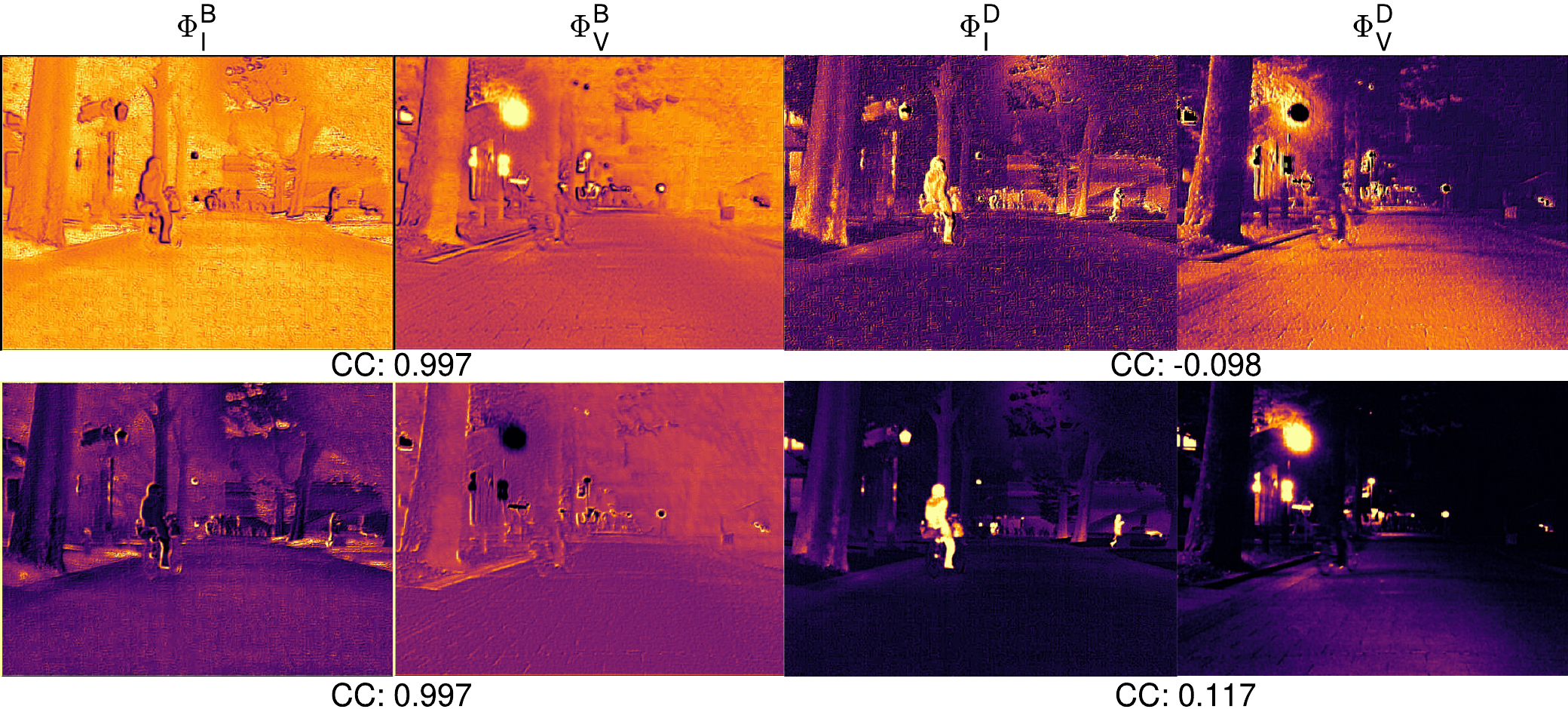}
    \caption{Visualization of the decomposed features}
    \label{fig:fvisual}
    \vspace{-1em}
\end{figure}

\bfsection{Visualization of feature decomposition}
\cref{fig:fvisual} visualizes the decomposed features. Obviously, more background information in the base feature group is activated, and the activated areas are also relevant. In the detail feature group, infrared features instead focus more on object highlights, while visible features pay more attention to details and textures, showing that the modality-specific features are well extracted. The visualization is consistent with our analysis.

\begin{table}[t]
    \centering
    \caption{Quantitative results of the IVF task. \textbf{Boldface} and {underline} show the best and second-best values, respectively.}
    \label{tab:Quantitative}%
    \resizebox{\linewidth}{!}{
        \begin{tabular}{ccccccccc}
            \toprule
            \multicolumn{9}{c}{\textbf{Dataset: MSRS Infrared-Visible Fusion Dataset}~\cite{DBLP:journals/inffus/TangYZJM22}}                                 \\
            &        EN        &        SD         &        SF         &        MI        &       SCD        &       VIF        &       Qbaf       &       SSIM       \\ \midrule
            DID~\cite{zhaoijcai2020} &       4.27       &       31.49       &       10.15       &       1.61       &       1.11       &       0.31       &       0.20       &       0.24       \\
            U2F~\cite{9151265}   &       5.37       &       25.52       &       9.07        &       1.40       &       1.24       &       0.54       &       0.42       &       0.77       \\
            SDN~\cite{DBLP:journals/ijcv/ZhangM21}  &       5.25       &       17.35       &       8.67        &       1.19       &       0.99       &       0.50       &       0.38       &       0.72       \\
            RFN~\cite{DBLP:conf/cvpr/Xu0YLL22}   &       5.56       &       24.09       &  \textbf{11.98}   &       1.30       &       1.13       &       0.51       &       0.43       &       0.83       \\
            TarD~\cite{DBLP:conf/cvpr/LiuFHWLZL22} &       5.28       &       25.22       &       5.98        &       1.49       &       0.71       &       0.42       &       0.18       &       0.47       \\
            DeF~\cite{Liang2022ECCV}  &       6.46       &       37.63       &       8.60        & \underline{2.16} &       1.35       & \underline{0.77} & \underline{0.54} & \underline{0.94} \\
            ReC~\cite{huangreconet}   & \underline{6.61} & \underline{43.24} &       9.77        &       2.16       & \underline{1.44} &       0.71       &       0.50       &       0.85       \\
            CDDFuse   &  \textbf{6.70}   &  \textbf{43.38}   & \underline{11.56} &  \textbf{3.47}   &  \textbf{1.62}   &  \textbf{1.05}   &  \textbf{0.69}   &  \textbf{1.00}   \\ \midrule
            \multicolumn{9}{c}{\textbf{Dataset: TNO Infrared-Visible Fusion Dataset}~\cite{TNO}}                                         \\
            &        EN        &        SD         &        SF         &        MI        &       SCD        &       VIF        &       Qbaf       &       SSIM       \\ \midrule
            DID~\cite{zhaoijcai2020} &       6.97       &       45.12       &       12.59       &       1.70       & \underline{1.71} & \underline{0.60} &       0.40       &       0.81       \\
            U2F~\cite{9151265} &       6.83       &       34.55       &       11.52       &       1.37       &       1.71       &       0.58       & \underline{0.44} &       0.99       \\
            SDN~\cite{DBLP:journals/ijcv/ZhangM21}  &       6.64       &       32.66       &       12.05       &       1.52       &       1.49       &       0.56       &       0.44       & \underline{1.00} \\
            RFN~\cite{DBLP:conf/cvpr/Xu0YLL22}  &       6.83       &       34.50       &  \textbf{15.71}   &       1.20       &       1.67       &       0.51       &       0.39       &       0.92       \\
            TarD~\cite{DBLP:conf/cvpr/LiuFHWLZL22} &       6.84       & \underline{45.63} &       8.68        & \underline{1.86} &       1.52       &       0.53       &       0.32       &       0.88       \\
            DeF~\cite{Liang2022ECCV} &       6.95       &       38.41       &       8.21        &       1.78       &       1.64       &       0.60       &       0.41       &       0.96       \\
            ReC~\cite{huangreconet}  & \underline{7.10} &       44.85       &       8.73        &       1.78       &       1.70       &       0.57       &       0.39       &       0.88       \\
            CDDFuse   &  \textbf{7.12}   &  \textbf{46.00}   & \underline{13.15} &  \textbf{2.19}   &  \textbf{1.76}   &  \textbf{0.77}   &  \textbf{0.54}   &  \textbf{1.03}   \\ \midrule
            \multicolumn{9}{c}{\textbf{Dataset: RoadScene Infrared-Visible Fusion Dataset}~\cite{xu2020aaai}}                                                \\
            &        EN        &        SD         &        SF         &        MI        &       SCD        &       VIF        &       Qbaf       &       SSIM       \\ \midrule
            DID~\cite{zhaoijcai2020} & \underline{7.43} &       51.58       &       14.66       &       2.11       &       1.70       &       0.58       &       0.48       &       0.86       \\
            U2F~\cite{9151265} &       7.09       &       38.12       &       13.25       &       1.87       &       1.70       &       0.60       & \underline{0.51} &       0.97       \\
            SDN~\cite{DBLP:journals/ijcv/ZhangM21}  &       7.14       &       40.20       &       13.70       &       2.21       &       1.49       &       0.60       &       0.51       &  \textbf{0.99}   \\
            RFN~\cite{DBLP:conf/cvpr/Xu0YLL22} &       7.21       &       41.25       & \underline{16.19} &       1.68       &       1.73       &       0.54       &       0.45       &       0.90       \\
            TarD~\cite{DBLP:conf/cvpr/LiuFHWLZL22} &       7.17       &       47.44       &       10.83       &       2.14       &       1.55       &       0.54       &       0.40       &       0.88       \\
            DeF~\cite{Liang2022ECCV}  &       7.23       &       44.44       &       10.22       & \underline{2.25} &       1.69       & \underline{0.63} &       0.48       &       0.89       \\
            ReC~\cite{huangreconet} &       7.36       & \underline{52.54} &       10.78       &       2.18       & \underline{1.74} &       0.59       &       0.43       &       0.88       \\
            CDDFuse   &  \textbf{7.44}   &  \textbf{54.67}   &  \textbf{16.36}   &  \textbf{2.30}   &  \textbf{1.81}   &  \textbf{0.69}   &  \textbf{0.52}   & \underline{0.98} \\ \bottomrule
    \end{tabular}}%
    \vspace{-1em}
\end{table}

\subsection{Ablation studies}\label{sec:Ablation}
Ablation experiments are set to verify the rationality of the different modules. EN, SD, VIF and SSIM are used to quantitatively validate the fusion effectiveness. The results of experimental groups are shown in Tab.~\ref{tab:Ablation}.

\bfsection{Decomposition loss $\mathcal{L}_{decomp}$}
In Exp.~\uppercase\expandafter{\romannumeral1}, we change the definition in Eq.~(\ref{equ:loss_d}) from division to subtraction as $\mathcal{L}_{decomp} = {(\mathcal{L}_{CC}^D)^2}-{\mathcal{L}_{CC}^B}$, which can also increase $\mathcal{L}_{CC}^B$ and decrease $(\mathcal{L}_{CC}^D)^2$.
The results of Exp.~\uppercase\expandafter{\romannumeral1} demonstrate that although the new loss can generate marginally satisfactory results, it produces poor results compared to the definition in Eq.~(\ref{equ:loss_d}).
In Exp.~\uppercase\expandafter{\romannumeral2}, we do not use the correlation-driven loss $\mathcal{L}_{decomp}$, and the results show that $\mathcal{L}_{decomp}$ is necessary for feature decomposition. There is no guarantee that BTE and DCE can learn the different frequency features without $\mathcal{L}_{decomp}$.

\bfsection{LT and the INN blocks}
We then verify the necessity of LT blocks and INN blocks in the \textit{Long-short Range Encoder}.
In Exp.~\uppercase\expandafter{\romannumeral3}, we changed the LT blocks as INN, \ie, the base and detail features are both extracted by INN blocks. Similarly, in Exp.~\uppercase\expandafter{\romannumeral4}, the features of different modalities are extracted by LT blocks.
The results show that although the ability of feature extraction for LT blocks is slightly stronger than that of INN blocks, it is worse than that of CDDFuse which cooperates with LT and INN blocks.
Subsequently, in Exp.~\uppercase\expandafter{\romannumeral5}, we changed the INN module as a CNN module composed of BRBs with similar parameters in INN blocks, and its effect is slightly worse than that of the LT module alone, which proves that the information loss is serious when the CNN is employed to accomplish the fusion task.

\bfsection{Two-stage training}
Finally, if we abandon the two-stage training and directly train the encoder, decoder and fusion layer simultaneously, the results are very unsatisfactory. It is proved that two-stage training can effectively reduce the difficulty of training and improve training robustness.

In summary, ablation results in Tab.~\ref{tab:Ablation} demonstrate the effectiveness and rationality of our network design.

\begin{table}[t]
    \centering
    \caption{Ablation experiment results in the testset of MSRS. \textbf{Bold} indicates the best value.}
    \label{tab:Ablation}
    \resizebox{\linewidth}{!}{
        \begin{tabular}{cccccc}
            \toprule
            &                       {Configurations}                       &      EN       &       SD       &      VIF      &     SSIM      \\ \midrule
            \uppercase\expandafter{\romannumeral1} & Division $\rightarrow$ Subtraction in $\mathcal{L}_{decomp}$ &     6.55      &     42.20      &     0.98      &     1.00      \\
            \uppercase\expandafter{\romannumeral2} &                  w/o $\mathcal{L}_{decomp}$                  &     6.19      &     36.49      &     0.96      &     0.97      \\
            \uppercase\expandafter{\romannumeral3} &              LT block $\rightarrow$ INN in BTE               &     6.47      &     41.39      &     1.00      &     0.98      \\
            \uppercase\expandafter{\romannumeral4} &              INN $\rightarrow$ LT block in DCE               &     6.56      &     42.18      &     1.00      &     0.99      \\
            \uppercase\expandafter{\romannumeral5} &             INN  $\rightarrow$ CNN block in DCE              &     6.54      &     42.10      &     0.98      &     0.98      \\
            \uppercase\expandafter{\romannumeral6} &                    w/o two-stage training                    &     6.28      &     38.42      &     0.97      &     0.99      \\ \midrule
            &                            {Ours}                            & \textbf{6.70} & \textbf{43.38} & \textbf{1.05} & \textbf{1.00} \\ \bottomrule
        \end{tabular}}
    \vspace{-1.5em}
\end{table}%

\subsection{Downstream IVF applications}
To further study fusion performance in high-level  MM computer vision tasks, this section applies the infrared, visible and fusion images of SOTA methods in \cref{sec:SOTA} to MM object detection and semantic segmentation, and investigate information fusion's benefit for the downstream tasks. Due to space constraints, qualitative results are presented in the supplementary material.

\subsubsection{Infrared-visible object detection}
\bfsection{Setup}
The MM object detection is performed on M$^3$FD dataset~\cite{DBLP:conf/cvpr/LiuFHWLZL22} with 4200 pairs of infrared/visible images, and six categories of labels (\ie, people, car, bus, motorcycle, truck and lamp). It is divided into training/validation/test sets with a proportion 8:1:1. YOLOv5~\cite{glenn_jocher_2020_4154370}, a SOTA detector, is employed to evaluate the detection performance with the metric mAP@0.5.
The training epoch, batch size, optimizer and initial learning rate are set as 400, 8, SGD optimizer and 1e-2, respectively.

\bfsection{Comparison with SOTA methods}
\cref{tab:MMOD} displays that CDDFuse has the best detection performance, especially in the People and Truck classes, demonstrating that CDDFuse can improve detection accuracy by fusing thermal radiation information and highlighting the difficult-to-observe targets.

\subsubsection{Infrared-visible semantic segmentation}
\bfsection{Setup} We conduct the MM semantic segmentation on the MSRS dataset~\cite{DBLP:journals/inffus/TangYZJM22} with semantic information of nine object categories (\ie, background, bump, color cone, guardrail, curve, bike, person, car stop and car). The division of dataset follows~\cite{DBLP:journals/inffus/TangYZJM22}. The backbone we choose is DeeplabV3+~\cite{DBLP:conf/eccv/ChenZPSA18} and we compare the model effectiveness by intersection-over-union (IoU).
All the models are supervised by cross-entropy loss and trained by SGD with the batchsize of 8 over 340 epochs, of which the first 100 epochs are trained by freezing the backbone network. The initial learning rate is 7e-3 and is decreased by the cosine annealing delay.

\bfsection{Comparison with SOTA methods}
The segmentation results are exhibited in~\cref{tab:MMSS}.
CDDFuse better integrates the edge and contour information in the source images, which enhances the ability of model to perceive the object boundary, and makes the segmentation more accurate.

\begin{table}[t]
    \centering
    \caption{AP@0.5(\%) values for MM detection on M$^3$FD dataset.}
    \label{tab:MMOD}%
    \resizebox{\linewidth}{!}{
        \begin{tabular}{cccccccc}
            \toprule
            &       Bus       &       Car       &       Lam       &       Mot       &       Peo       &       Tru       &     mAP@0.5     \\ \midrule
            IR  &      78.75      &      88.69      &      70.17      &      63.42      &      80.91      &      65.77      &      74.62      \\
            VI  &      78.29      &      90.73      &      86.35      &      69.33      &      70.53      &      70.91      &      77.69      \\
            DID  &      79.65      &      92.51      &      84.70      &      68.72      &      79.61      &      68.78      &      78.99      \\
            U2F  &      79.15      &      92.29      & \underline{87.61} &      66.75      &      80.67      &      71.37      &      79.64      \\
            SDN  &      81.44      &      92.33      &      84.14      &      67.37      &      79.35      &      69.29      &      78.99      \\
            RFN  &      78.15      &      91.94      &      84.95      & \textbf{72.80} &      79.41      &      69.04      &      79.38      \\
            TarD &      81.33      & \textbf{94.76} &      87.13      &      69.34      & \underline{81.52} &      68.65      &      80.45      \\
            DeF  & \textbf{82.94} &      92.49      & \textbf{87.78} &      69.45      &      80.82      & \underline{71.44} & \underline{80.82} \\
            ReC  &      78.92      &      91.79      &      87.41      &      69.34      &      79.41      &      69.98      &      79.48      \\
            Ours & \underline{82.60} & \underline{92.54} &      86.88      & \underline{71.62} & \textbf{81.60} & \textbf{71.53} & \textbf{81.13}\\ \bottomrule
    \end{tabular}}
\end{table}%

\begin{table}[t]
    \centering
    \caption{IoU(\%) values for MM segmentation on MSRS dataset.}
    \label{tab:MMSS}
    \resizebox{\linewidth}{!}{
        \begin{tabular}{ccccccccccc}
            \toprule
            Models                 &        Unl        &        Car        &        Per        &        Bik        &        Cur        &        CS         &        GD         &        CC         &        Bu         &       mIOU        \\ \midrule
            VI                   &       90.5        &       75.6        &       45.4        &       59.4        &       37.2        &       51.0        &       46.4        &       43.5        &       50.2        &       55.4        \\
            IR                   &       84.7        &       67.8        &       56.4        &       51.8        &       34.6        &       39.3        &       42.2        &       40.2        &       48.4        &       51.7        \\
            DID~\cite{zhaoijcai2020}        &       97.2        &       78.3        &       58.7        &       60.9        &       36.2        & \underline{52.9} &  \textbf{62.4}   &       44.0        &       55.7        &       60.7        \\
            U2F~\cite{9151265}           &       97.5        &       82.3        & \underline{63.4} &       62.6        &       40.3        &       52.6        &       51.9        &       44.8        &  \textbf{59.5}   & \underline{61.7} \\
            SDN~\cite{DBLP:journals/ijcv/ZhangM21} &       97.3        &       78.4        &       62.5        &       61.7        &       35.7        &       49.3        &       52.4        &       42.2        &       52.9        &       59.2        \\
            RFN~\cite{DBLP:conf/cvpr/Xu0YLL22}   &       97.3        &       78.7        &       60.6        &       61.3        &       36.3        &       49.4        &       45.6        &       45.7        &       48.0        &       58.1        \\
            TarD~\cite{DBLP:conf/cvpr/LiuFHWLZL22} &       97.1        &       79.1        &       55.4        &       59.0        &       33.6        &       49.4        &       54.9        &       42.6        &       53.5        &       58.3        \\
            DeF~\cite{Liang2022ECCV}        & \underline{97.5} & \underline{82.6} &       61.1        & \underline{62.6} &       40.4        &       51.5        &       48.1        & \underline{47.9} &       54.8        &       60.7        \\
            ReC~\cite{huangreconet}         &       97.4        &       81.0        &       59.9        &       61.4        & \underline{41.0} &       51.3        &       54.4        &       47.4        &       55.9        &       61.1        \\
            Ours                  &  \textbf{97.7}   &  \textbf{84.6}   &  \textbf{64.2}   &  \textbf{65.1}   &  \textbf{43.9}   &  \textbf{53.8}   & \underline{61.7} &  \textbf{50.6}   & \underline{57.3} &  \textbf{64.3}   \\ \bottomrule
    \end{tabular}}
\end{table}%

\begin{figure}[t]
    \centering
    \includegraphics[width=\linewidth]{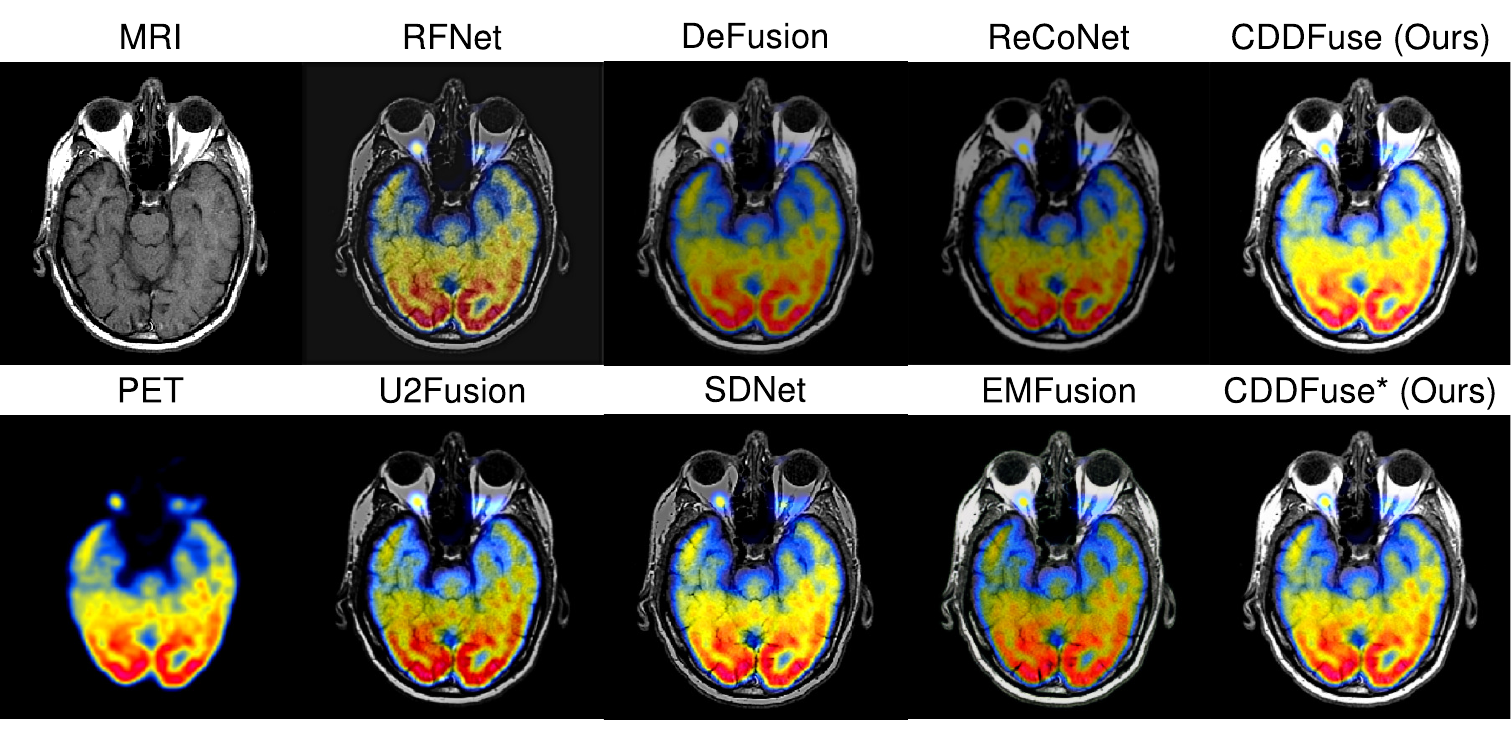}
    \caption{Visual comparison for ``MRI-PET-16'' in MRI-PET MIF.}
    \label{fig:MIF1}
    \vspace{-1em}
\end{figure}

\begin{table}[t]
    \centering
    \caption{Quantitative results of the MIF task. \textbf{Boldface} and {underline} show the best and second-best values, respectively. CDDFuse$^*$ represents the results after training on MIF datasets.}
    \label{tab:MIF}%
    \resizebox{\linewidth}{!}{
        \begin{tabular}{ccccccccc}
            \toprule
            \multicolumn{9}{c}{\textbf{Dataset: MRI-CT Medical Image Fusion}}                                                                              \\
            &        EN        &        SD         &        SF         &        MI        &       SCD        &       VIF       &    $Q^{AB/F}$    &       SSIM       \\ \midrule
            TarD~\cite{DBLP:conf/cvpr/LiuFHWLZL22}    &       4.75       &       61.14       &       28.38       &       1.94       &       0.81       &       0.32       &       0.35       &       0.61       \\
            RFN~\cite{DBLP:conf/cvpr/Xu0YLL22}      &  \textbf{5.30}   &       52.95       & \underline{33.42} &       1.98       &       0.58       &       0.33       & \underline{0.52} &       0.49       \\
            DeF~\cite{Liang2022ECCV}           &       4.63       &       66.38       &       21.56       & \underline{2.20} &       1.12       & \underline{0.47} &       0.44       & \underline{1.29} \\
            ReC~\cite{huangreconet}           &       4.41       & \underline{66.96} &       20.16       &       2.03       & \underline{1.24} &       0.40       &       0.42       &       1.29       \\
            CDDFuse                   & \underline{4.83} &  \textbf{88.59}   &  \textbf{33.83}   &  \textbf{2.24}   &  \textbf{1.74}   &  \textbf{0.50}   &  \textbf{0.59}   &  \textbf{1.31}   \\
            \cdashline{1-9}[1pt/1pt]
            U2F~\cite{9151265} &       4.88       &       52.98       &       22.54       &       2.08       &       0.75       &       0.37       &       0.46       &       0.49       \\
            SDN~\cite{DBLP:journals/ijcv/ZhangM21}    &  \textbf{5.02}   &       60.07       & \underline{29.41} &       2.14       &       0.97       &       0.38       &       0.47       &       0.51       \\
            EMF~\cite{DBLP:journals/inffus/Xu021}    &       4.76       & \underline{72.76} &       22.56       & \underline{2.34} & \underline{1.32} & \underline{0.56} & \underline{0.49} & \underline{1.31} \\
            CDDFuse$^*$                  & \underline{4.88} &  \textbf{79.17}   &  \textbf{38.14}   &  \textbf{2.61}   &  \textbf{1.41}   &  \textbf{0.61}   &  \textbf{0.68}   &  \textbf{1.34}   \\ \midrule
            \multicolumn{9}{c}{\textbf{Dataset: MRI-PET Medical Image Fusion}}                                                                             \\
            &        EN        &        SD         &        SF         &        MI        &       SCD        &       VIF       &    $Q^{AB/F}$    &       SSIM       \\ \midrule
            TarD~\cite{DBLP:conf/cvpr/LiuFHWLZL22}    &       3.81       &       57.65       &       23.65       &       1.36       &       1.46       &       0.57       & \underline{0.58} &       0.68       \\
            RFN~\cite{DBLP:conf/cvpr/Xu0YLL22}      &  \textbf{4.77}   &       50.57       &  \textbf{29.11}   &       1.53       &       0.96       &       0.39       &       0.52       &       0.42       \\
            DeF~\cite{Liang2022ECCV}           &       4.17       &       64.65       &       22.35       & \underline{1.74} &       1.48       & \underline{0.58} &       0.56       & \underline{1.45} \\
            ReC~\cite{huangreconet}           &       3.66       & \underline{65.25} &       21.72       &       1.51       & \underline{1.49} &       0.44       &       0.51       &       1.40       \\
            CDDFuse                   & \underline{4.24} &  \textbf{81.72}   & \underline{28.04} &  \textbf{1.87}   &  \textbf{1.82}   &  \textbf{0.66}   &  \textbf{0.65}   &  \textbf{1.46}   \\
            \cdashline{1-9}[1pt/1pt]
            U2F~\cite{9151265} &       3.73       &       57.07       &       23.27       &       1.69       &       1.27       &       0.40       &       0.49       &       1.39       \\
            SDN~\cite{DBLP:journals/ijcv/ZhangM21}    &       3.83       & \underline{61.40} &  \textbf{31.97}   &       1.71       & \underline{1.40} &       0.47       &       0.57       &       1.46       \\
            EMF~\cite{DBLP:journals/inffus/Xu021}    & \underline{4.21} &       56.80       &       26.01       & \underline{1.82} &       1.31       & \underline{0.62} & \underline{0.67} & \underline{1.47} \\
            CDDFuse$^*$                   &  \textbf{4.23}   &  \textbf{70.73}   & \underline{29.57} &  \textbf{2.03}   &  \textbf{1.69}   &  \textbf{0.71}   &  \textbf{0.71}   &  \textbf{1.49}   \\ \midrule
            \multicolumn{9}{c}{\textbf{Dataset: MRI-SPECT Medical Image Fusion}}                                                                            \\
            &        EN        &        SD         &        SF         &        MI        &       SCD        &       VIF       &    $Q^{AB/F}$    &       SSIM       \\ \midrule
            TarD~\cite{DBLP:conf/cvpr/LiuFHWLZL22}    &       3.66       &       53.46       &       18.50       &       1.44       &       0.90       & \underline{0.64} &       0.52       &       0.36       \\
            RFN~\cite{DBLP:conf/cvpr/Xu0YLL22}      &  \textbf{4.39}   &       44.01       &  \textbf{23.77}   &       1.60       &       0.72       &       0.45       & \underline{0.58} &       0.37       \\
            DeF~\cite{Liang2022ECCV}           &       3.81       &       56.65       &       15.45       & \underline{1.80} &       1.27       &       0.61       &       0.56       &  \textbf{1.46}   \\
            ReC~\cite{huangreconet}           &       3.22       & \underline{60.07} &       17.40       &       1.50       & \underline{1.47} &       0.46       &       0.54       &       1.40       \\
            CDDFuse                   & \underline{3.91} &  \textbf{71.82}   & \underline{20.68} &  \textbf{1.89}   &  \textbf{1.92}   &  \textbf{0.66}   &  \textbf{0.69}   & \underline{1.44} \\
            \cdashline{1-9}[1pt/1pt]
            U2F~\cite{9151265} &       3.47       & \underline{52.97} &       19.58       &       1.68       & \underline{1.28} &       0.48       &       0.57       &       1.41       \\
            SDN~\cite{DBLP:journals/ijcv/ZhangM21}    &       3.43       &       49.62       &  \textbf{22.20}   &       1.69       &       1.09       &       0.55       &       0.66       &       1.48       \\
            EMF~\cite{DBLP:journals/inffus/Xu021}    & \underline{3.74} &       51.93       &       17.14       & \underline{1.88} &       1.12       & \underline{0.71} & \underline{0.74} &  \textbf{1.49}   \\
            CDDFuse$^*$                   &  \textbf{3.90}   &  \textbf{58.31}   & \underline{20.87} &  \textbf{2.49}   &  \textbf{1.35}   &  \textbf{0.97}   &  \textbf{0.78}   & \underline{1.48} \\ \bottomrule
    \end{tabular}}
    \vspace{-1em}
\end{table}%

\section{Medical image fusion}

\bfsection{Setup} We selected 286 pairs of medical images from the Harvard Medical website~\cite{HarvardMedical} for MIF experiments, of which 130 pairs and 20 pairs are used for training and validation, respectively. 21 pairs of MRI-CT images, 42 pairs of MRI-PET images and 73 pairs of MRI-SPECT images are utilized as the test datasets. The training strategy and the metrics for evaluation are the same as that for IVF.

\bfsection{Comparison methods}
We performed two groups of experiments. First, we compare fusion methods trained on the IVF task, \ie TarDAL~\cite{DBLP:conf/cvpr/LiuFHWLZL22}, RFNet~\cite{DBLP:conf/cvpr/Xu0YLL22}, DeFusion~\cite{Liang2022ECCV}, ReCoNet~\cite{huangreconet} and our CDDFuse, to demonstrate the generalization ability of the fusion methods. Note that the above methods are not fine-tuned on the MIF dataset.
Then, we train CDDFuse on the MIF dataset (denoted as CDDFuse$^*$), and compare it with U2Fusion~\cite{9151265}, SDNet~\cite{DBLP:journals/ijcv/ZhangM21} and EMFusion~\cite{DBLP:journals/inffus/Xu021}, which are all trained on the MIF dataset.

\bfsection{Comparison with SOTA methods}
Qualitative and Quantitative results are presented in \cref{fig:MIF1,tab:MIF}.
CDDFuse can preserve the detailed texture and highlight the structure information, and achieves leading performance on almost all metrics whether trained on the MIF dataset or not.
\section{Conclusion}\label{sec:5}
In this paper, we propose a dual-branch Transformer-CNN architecture for multi-modal image fusion. With the help of Restormer, Lite transformer and invertible neural network blocks, modality-specific and -shared features are better extracted, and the decomposition for them is more intuitive and effective by the proposed correlation-driven decomposition loss. Experiments demonstrate the fusion effect of our CDDFuse, and the accuracy of downstream multi-modal pattern recognition tasks can be also improved.

\section*{Acknowledgement}\label{sec:6}
This work has been supported by the National Key Research and Development Program of China under grant 2018AAA0102201, the National Natural Science Foundation of China under Grant 61976174 and 12201497,
the Guangdong Basic and Applied Basic Research Foundation under Grant 2023A1515011358, the Fundamental Research Funds for the Central Universities under Grant D5000220060, and partly supported by the Alexander von Humboldt Foundation.

{\small
    \bibliographystyle{ieee_fullname}
    \bibliography{xbib}

\begin{thebibliography}{10}\itemsep=-1pt

\bibitem{HarvardMedical}
Harvard medical website.
\newblock \url{http://www.med.harvard.edu/AANLIB/home.html}.

\bibitem{DBLP:conf/iclr/ArdizzoneKRK19}
Lynton Ardizzone, Jakob Kruse, Carsten Rother, and Ullrich K{\"{o}}the.
\newblock Analyzing inverse problems with invertible neural networks.
\newblock In {\em {ICLR}}, 2019.

\bibitem{DBLP:journals/corr/abs-1907-02392}
Lynton Ardizzone, Carsten L{\"{u}}th, Jakob Kruse, Carsten Rother, and Ullrich
  K{\"{o}}the.
\newblock Guided image generation with conditional invertible neural networks.
\newblock {\em CoRR}, abs/1907.02392, 2019.

\bibitem{DBLP:conf/cvpr/BandaraP22}
Wele Gedara~Chaminda Bandara and Vishal~M. Patel.
\newblock Hypertransformer: {A} textural and spectral feature fusion
  transformer for pansharpening.
\newblock In {\em {CVPR}}, pages 1757--1767, 2022.

\bibitem{DBLP:conf/icml/BehrmannGCDJ19}
Jens Behrmann, Will Grathwohl, Ricky T.~Q. Chen, David Duvenaud, and
  J{\"{o}}rn{-}Henrik Jacobsen.
\newblock Invertible residual networks.
\newblock In {\em {ICML}}, volume~97, pages 573--582, 2019.

\bibitem{DBLP:journals/corr/abs-2004-10934}
Alexey Bochkovskiy, Chien{-}Yao Wang, and Hong{-}Yuan~Mark Liao.
\newblock Yolov4: Optimal speed and accuracy of object detection.
\newblock {\em CoRR}, abs/2004.10934, 2020.

\bibitem{DBLP:journals/corr/abs-2106-06847}
Jiezhang Cao, Yawei Li, Kai Zhang, and Luc~Van Gool.
\newblock Video super-resolution transformer.
\newblock {\em CoRR}, abs/2106.06847, 2021.

\bibitem{DBLP:conf/eccv/CarionMSUKZ20}
Nicolas Carion, Francisco Massa, Gabriel Synnaeve, Nicolas Usunier, Alexander
  Kirillov, and Sergey Zagoruyko.
\newblock End-to-end object detection with transformers.
\newblock In {\em {ECCV}}, pages 213--229, 2020.

\bibitem{DBLP:conf/cvpr/Chen000DLMX0021}
Hanting Chen, Yunhe Wang, Tianyu Guo, Chang Xu, Yiping Deng, Zhenhua Liu, Siwei
  Ma, Chunjing Xu, Chao Xu, and Wen Gao.
\newblock Pre-trained image processing transformer.
\newblock In {\em {CVPR}}, pages 12299--12310, 2021.

\bibitem{DBLP:conf/eccv/ChenZPSA18}
Liang{-}Chieh Chen, Yukun Zhu, George Papandreou, Florian Schroff, and Hartwig
  Adam.
\newblock Encoder-decoder with atrous separable convolution for semantic image
  segmentation.
\newblock In {\em {ECCV}}, pages 833--851, 2018.

\bibitem{DBLP:journals/pami/0002D21}
Xin Deng and Pier~Luigi Dragotti.
\newblock Deep convolutional neural network for multi-modal image restoration
  and fusion.
\newblock {\em {IEEE} TPAMI}, 43(10):3333--3348, 2021.

\bibitem{DBLP:journals/corr/DinhKB14}
Laurent Dinh, David Krueger, and Yoshua Bengio.
\newblock {NICE:} non-linear independent components estimation.
\newblock In {\em {ICLR} (Workshop)}, 2015.

\bibitem{DBLP:conf/iclr/DinhSB17}
Laurent Dinh, Jascha Sohl{-}Dickstein, and Samy Bengio.
\newblock Density estimation using real {NVP}.
\newblock In {\em {ICLR}}, 2017.

\bibitem{DBLP:conf/iclr/DosovitskiyB0WZ21}
Alexey Dosovitskiy, Lucas Beyer, Alexander Kolesnikov, Dirk Weissenborn,
  Xiaohua Zhai, Thomas Unterthiner, Mostafa Dehghani, Matthias Minderer, Georg
  Heigold, Sylvain Gelly, Jakob Uszkoreit, and Neil Houlsby.
\newblock An image is worth 16x16 words: Transformers for image recognition at
  scale.
\newblock In {\em {ICLR}}, 2021.

\bibitem{DBLP:journals/tip/GaoDXXD22}
Fangyuan Gao, Xin Deng, Mai Xu, Jingyi Xu, and Pier~Luigi Dragotti.
\newblock Multi-modal convolutional dictionary learning.
\newblock {\em {IEEE} TIP}, 31:1325--1339, 2022.

\bibitem{DBLP:journals/corr/abs-2203-14186}
Zhicheng Geng, Luming Liang, Tianyu Ding, and Ilya Zharkov.
\newblock {RSTT:} real-time spatial temporal transformer for space-time video
  super-resolution.
\newblock {\em CoRR}, abs/2203.14186, 2022.

\bibitem{DBLP:conf/nips/GomezRUG17}
Aidan~N. Gomez, Mengye Ren, Raquel Urtasun, and Roger~B. Grosse.
\newblock The reversible residual network: Backpropagation without storing
  activations.
\newblock In {\em {NIPS}}, pages 2214--2224, 2017.

\bibitem{DBLP:conf/nips/GoodfellowPMXWOCB14}
Ian~J. Goodfellow, Jean Pouget{-}Abadie, Mehdi Mirza, Bing Xu, David
  Warde{-}Farley, Sherjil Ozair, Aaron~C. Courville, and Yoshua Bengio.
\newblock Generative adversarial nets.
\newblock In {\em {NIPS}}, pages 2672--2680, 2014.

\bibitem{huangreconet}
Zhanbo Huang, Jinyuan Liu, Xin Fan, Risheng Liu, Wei Zhong, and Zhongxuan Luo.
\newblock Reconet: Recurrent correction network for fast and efficient
  multi-modality image fusion.
\newblock In {\em {ECCV}}, 2022.

\bibitem{DBLP:conf/iclr/JacobsenSO18}
J{\"{o}}rn{-}Henrik Jacobsen, Arnold W.~M. Smeulders, and Edouard Oyallon.
\newblock i-revnet: Deep invertible networks.
\newblock In {\em {ICLR}}, 2018.

\bibitem{DBLP:journals/inffus/JamesD14}
Alex~Pappachen James and Belur~V. Dasarathy.
\newblock Medical image fusion: {A} survey of the state of the art.
\newblock {\em Inf. Fusion}, 19:4--19, 2014.

\bibitem{DBLP:conf/mm/JiangZ0L22}
Zhiying Jiang, Zengxi Zhang, Xin Fan, and Risheng Liu.
\newblock Towards all weather and unobstructed multi-spectral image stitching:
  Algorithm and benchmark.
\newblock In {\em {ACM} Multimedia}, pages 3783--3791, 2022.

\bibitem{DBLP:conf/iccv/Jing0XWG21}
Junpeng Jing, Xin Deng, Mai Xu, Jianyi Wang, and Zhenyu Guan.
\newblock Hinet: Deep image hiding by invertible network.
\newblock In {\em {ICCV}}, pages 4713--4722, 2021.

\bibitem{glenn_jocher_2020_4154370}
Glenn Jocher.
\newblock {ultralytics/yolov5}.
\newblock \url{https://github.com/ultralytics/yolov5}, oct 2020.

\bibitem{DBLP:conf/mm/JuZLZ20}
Xincheng Ju, Dong Zhang, Junhui Li, and Guodong Zhou.
\newblock Transformer-based label set generation for multi-modal multi-label
  emotion detection.
\newblock In {\em {ACM} Multimedia}, pages 512--520, 2020.

\bibitem{DBLP:journals/tip/JungKJHS20}
Hyungjoo Jung, Youngjung Kim, Hyunsung Jang, Namkoo Ha, and Kwanghoon Sohn.
\newblock Unsupervised deep image fusion with structure tensor representations.
\newblock {\em {IEEE} TIP}, 29:3845--3858, 2020.

\bibitem{DBLP:conf/nips/KingmaD18}
Diederik~P. Kingma and Prafulla Dhariwal.
\newblock Glow: Generative flow with invertible 1x1 convolutions.
\newblock In {\em NeurIPS}, pages 10236--10245, 2018.

\bibitem{DBLP:journals/inffus/LiWK21}
Hui Li, Xiao{-}Jun Wu, and Josef Kittler.
\newblock Rfn-nest: An end-to-end residual fusion network for infrared and
  visible images.
\newblock {\em Inf. Fusion}, 73:72--86, 2021.

\bibitem{li2018densefuse}
Hui Li and Xiao-Jun Wu.
\newblock Densefuse: A fusion approach to infrared and visible images.
\newblock {\em IEEE TIP}, 28(5):2614--2623, 2018.

\bibitem{DBLP:journals/corr/abs-2112-10175}
Wenbo Li, Xin Lu, Jiangbo Lu, Xiangyu Zhang, and Jiaya Jia.
\newblock On efficient transformer and image pre-training for low-level vision.
\newblock {\em CoRR}, abs/2112.10175, 2021.

\bibitem{DBLP:conf/iccvw/LiangCSZGT21}
Jingyun Liang, Jiezhang Cao, Guolei Sun, Kai Zhang, Luc~Van Gool, and Radu
  Timofte.
\newblock Swinir: Image restoration using swin transformer.
\newblock In {\em {ICCVW}}, pages 1833--1844, 2021.

\bibitem{Liang2022ECCV}
Pengwei Liang, Junjun Jiang, Xianming Liu, and Jiayi Ma.
\newblock Fusion from decomposition: A self-supervised decomposition approach
  for image fusion.
\newblock In {\em {ECCV}}, 2022.

\bibitem{Liu2019Perceptual}
Aishan Liu, Xianglong Liu, Jiaxin Fan, Yuqing Ma, Anlan Zhang, Huiyuan Xie, and
  Dacheng Tao.
\newblock Perceptual-sensitive {GAN} for generating adversarial patches.
\newblock In {\em {AAAI}}, pages 1028--1035, 2019.

\bibitem{liu2021ANP}
Aishan Liu, Xianglong Liu, Hang Yu, Chongzhi Zhang, Qiang Liu, and Dacheng Tao.
\newblock Training robust deep neural networks via adversarial noise
  propagation.
\newblock {\em IEEE TIP}, 2021.

\bibitem{DBLP:conf/cvpr/LiuFHWLZL22}
Jinyuan Liu, Xin Fan, Zhanbo Huang, Guanyao Wu, Risheng Liu, Wei Zhong, and
  Zhongxuan Luo.
\newblock Target-aware dual adversarial learning and a multi-scenario
  multi-modality benchmark to fuse infrared and visible for object detection.
\newblock In {\em {CVPR}}, pages 5792--5801, 2022.

\bibitem{DBLP:journals/tcsv/LiuFJLL22}
Jinyuan Liu, Xin Fan, Ji Jiang, Risheng Liu, and Zhongxuan Luo.
\newblock Learning a deep multi-scale feature ensemble and an edge-attention
  guidance for image fusion.
\newblock {\em {IEEE} TCSVT}, 32(1):105--119, 2022.

\bibitem{DBLP:conf/mm/LiuLL021}
Risheng Liu, Zhu Liu, Jinyuan Liu, and Xin Fan.
\newblock Searching a hierarchically aggregated fusion architecture for fast
  multi-modality image fusion.
\newblock In {\em {ACM} Multimedia}, pages 1600--1608, 2021.

\bibitem{DBLP:conf/iccv/LiuL00W0LG21}
Ze Liu, Yutong Lin, Yue Cao, Han Hu, Yixuan Wei, Zheng Zhang, Stephen Lin, and
  Baining Guo.
\newblock Swin transformer: Hierarchical vision transformer using shifted
  windows.
\newblock In {\em {ICCV}}, pages 9992--10002, 2021.

\bibitem{ma2020infrared}
Jiayi Ma, Pengwei Liang, Wei Yu, Chen Chen, Xiaojie Guo, Jia Wu, and Junjun
  Jiang.
\newblock Infrared and visible image fusion via detail preserving adversarial
  learning.
\newblock {\em Inf. Fusion}, 54:85--98, 2020.

\bibitem{ma2019infrared}
Jiayi Ma, Yong Ma, and Chang Li.
\newblock Infrared and visible image fusion methods and applications: A survey.
\newblock {\em Inf. Fusion}, 45:153--178, 2019.

\bibitem{DBLP:journals/ieeejas/MaTFHMM22}
Jiayi Ma, Linfeng Tang, Fan Fan, Jun Huang, Xiaoguang Mei, and Yong Ma.
\newblock Swinfusion: Cross-domain long-range learning for general image fusion
  via swin transformer.
\newblock {\em {IEEE} {CAA} J. Autom. Sinica}, 9(7):1200--1217, 2022.

\bibitem{ma2019fusiongan}
Jiayi Ma, Wei Yu, Pengwei Liang, Chang Li, and Junjun Jiang.
\newblock Fusiongan: A generative adversarial network for infrared and visible
  image fusion.
\newblock {\em Inf. Fusion}, 48:11--26, 2019.

\bibitem{DBLP:journals/tim/MaZSLX21}
Jiayi Ma, Hao Zhang, Zhenfeng Shao, Pengwei Liang, and Han Xu.
\newblock Ganmcc: {A} generative adversarial network with multiclassification
  constraints for infrared and visible image fusion.
\newblock {\em {IEEE} TIM}, 70:1--14, 2021.

\bibitem{DBLP:journals/ijcv/MaZJZG19}
Jiayi Ma, Ji Zhao, Junjun Jiang, Huabing Zhou, and Xiaojie Guo.
\newblock Locality preserving matching.
\newblock {\em Int. J. Comput. Vis.}, 127(5):512--531, 2019.

\bibitem{MEF-SSIM2018TCI}
Kede Ma, Zhengfang Duanmu, Hojatollah Yeganeh, and Zhou Wang.
\newblock Multi-exposure image fusion by optimizing {A} structural similarity
  index.
\newblock {\em {IEEE} TCI}, 4(1):60--72, 2018.

\bibitem{mao2017least}
Xudong Mao, Qing Li, Haoran Xie, Raymond~YK Lau, Zhen Wang, and Stephen
  Paul~Smolley.
\newblock Least squares generative adversarial networks.
\newblock In {\em {CVPR}}, pages 2794--2802, 2017.

\bibitem{meher2019a}
Bikash Meher, Sanjay Agrawal, Rutuparna Panda, and Ajith Abraham.
\newblock A survey on region based image fusion methods.
\newblock {\em Inf. Fusion}, 48:119--132, 2019.

\bibitem{mirza2014conditional}
Mehdi Mirza and Simon Osindero.
\newblock Conditional generative adversarial nets.
\newblock {\em arXiv preprint arXiv:1411.1784}, 2014.

\bibitem{qin2022bibert}
Haotong Qin, Yifu Ding, Mingyuan Zhang, YAN Qinghua, Aishan Liu, Qingqing Dang,
  Ziwei Liu, and Xianglong Liu.
\newblock Bibert: Accurate fully binarized bert.
\newblock In {\em {ICLR}}, 2022.

\bibitem{qin2020forward}
Haotong Qin, Ruihao Gong, Xianglong Liu, Mingzhu Shen, Ziran Wei, Fengwei Yu,
  and Jingkuan Song.
\newblock Forward and backward information retention for accurate binary neural
  networks.
\newblock In {\em {CVPR}}, pages 2250--2259, 2020.

\bibitem{qin2022distribution}
Haotong Qin, Xiangguo Zhang, Ruihao Gong, Yifu Ding, Yi Xu, and Xianglong Liu.
\newblock Distribution-sensitive information retention for accurate binary
  neural network.
\newblock {\em International Journal of Computer Vision}, pages 1--22, 2022.

\bibitem{DBLP:conf/cvpr/QinZHGDJ19}
Xuebin Qin, Zichen~Vincent Zhang, Chenyang Huang, Chao Gao, Masood Dehghan, and
  Martin J{\"{a}}gersand.
\newblock Basnet: Boundary-aware salient object detection.
\newblock In {\em {CVPR}}, pages 7479--7489, 2019.

\bibitem{DBLP:conf/cvpr/SandlerHZZC18}
Mark Sandler, Andrew~G. Howard, Menglong Zhu, Andrey Zhmoginov, and
  Liang{-}Chieh Chen.
\newblock Mobilenetv2: Inverted residuals and linear bottlenecks.
\newblock In {\em {CVPR}}, pages 4510--4520, 2018.

\bibitem{DBLP:conf/mm/SunCZH22}
Yiming Sun, Bing Cao, Pengfei Zhu, and Qinghua Hu.
\newblock Detfusion: {A} detection-driven infrared and visible image fusion
  network.
\newblock In {\em {ACM} Multimedia}, pages 4003--4011, 2022.

\bibitem{DBLP:journals/ieeejas/TangDMHM22}
Linfeng Tang, Yuxin Deng, Yong Ma, Jun Huang, and Jiayi Ma.
\newblock Superfusion: {A} versatile image registration and fusion network with
  semantic awareness.
\newblock {\em {IEEE} {CAA} J. Autom. Sinica}, 9(12):2121--2137, 2022.

\bibitem{DBLP:journals/inffus/TangYM22}
Linfeng Tang, Jiteng Yuan, and Jiayi Ma.
\newblock Image fusion in the loop of high-level vision tasks: {A}
  semantic-aware real-time infrared and visible image fusion network.
\newblock {\em Inf. Fusion}, 82:28--42, 2022.

\bibitem{DBLP:journals/inffus/TangYZJM22}
Linfeng Tang, Jiteng Yuan, Hao Zhang, Xingyu Jiang, and Jiayi Ma.
\newblock Piafusion: {A} progressive infrared and visible image fusion network
  based on illumination aware.
\newblock {\em Inf. Fusion}, 83-84:79--92, 2022.

\bibitem{tang2021robustart}
Shiyu Tang, Ruihao Gong, Yan Wang, Aishan Liu, Jiakai Wang, Xinyun Chen,
  Fengwei Yu, Xianglong Liu, Dawn Song, Alan~L. Yuille, Philip H.~S. Torr, and
  Dacheng Tao.
\newblock Robustart: Benchmarking robustness on architecture design and
  training techniques.
\newblock {\em CoRR}, abs/2109.05211, 2021.

\bibitem{TNO}
Alexander Toet and Maarten~A. Hogervorst.
\newblock {Progress in color night vision}.
\newblock {\em Optical Engineering}, 51(1):1 -- 20, 2012.

\bibitem{DBLP:conf/icml/TouvronCDMSJ21}
Hugo Touvron, Matthieu Cord, Matthijs Douze, Francisco Massa, Alexandre
  Sablayrolles, and Herv{\'{e}} J{\'{e}}gou.
\newblock Training data-efficient image transformers {\&} distillation through
  attention.
\newblock In {\em {ICML}}, volume 139, pages 10347--10357, 2021.

\bibitem{DBLP:conf/nips/VaswaniSPUJGKP17}
Ashish Vaswani, Noam Shazeer, Niki Parmar, Jakob Uszkoreit, Llion Jones,
  Aidan~N. Gomez, Lukasz Kaiser, and Illia Polosukhin.
\newblock Attention is all you need.
\newblock In {\em {NIPS}}, pages 5998--6008, 2017.

\bibitem{DBLP:conf/ijcai/WangLFL22}
Di Wang, Jinyuan Liu, Xin Fan, and Risheng Liu.
\newblock Unsupervised misaligned infrared and visible image fusion via
  cross-modality image generation and registration.
\newblock In {\em {IJCAI}}, pages 3508--3515, 2022.

\bibitem{Wang2021DualAttention}
Jiakai Wang, Aishan Liu, Zixin Yin, Shunchang Liu, Shiyu Tang, and Xianglong
  Liu.
\newblock Dual attention suppression attack: Generate adversarial camouflage in
  physical world.
\newblock In {\em {CVPR}}, 2021.

\bibitem{DBLP:conf/iccv/WangX0FSLL0021}
Wenhai Wang, Enze Xie, Xiang Li, Deng{-}Ping Fan, Kaitao Song, Ding Liang, Tong
  Lu, Ping Luo, and Ling Shao.
\newblock Pyramid vision transformer: {A} versatile backbone for dense
  prediction without convolutions.
\newblock In {\em {ICCV}}, pages 548--558, 2021.

\bibitem{wang2004image}
Zhou Wang, Alan~C Bovik, Hamid~R Sheikh, Eero~P Simoncelli, et~al.
\newblock Image quality assessment: from error visibility to structural
  similarity.
\newblock {\em IEEE TIP}, 13(4):600--612, 2004.

\bibitem{DBLP:journals/corr/abs-2106-03106}
Zhendong Wang, Xiaodong Cun, Jianmin Bao, and Jianzhuang Liu.
\newblock Uformer: {A} general u-shaped transformer for image restoration.
\newblock {\em CoRR}, abs/2106.03106, 2021.

\bibitem{DBLP:conf/iclr/WuLLLH20}
Zhanghao Wu, Zhijian Liu, Ji Lin, Yujun Lin, and Song Han.
\newblock Lite transformer with long-short range attention.
\newblock In {\em {ICLR}}, 2020.

\bibitem{DBLP:conf/eccv/XiaoZLWHKBLL20}
Mingqing Xiao, Shuxin Zheng, Chang Liu, Yaolong Wang, Di He, Guolin Ke, Jiang
  Bian, Zhouchen Lin, and Tie{-}Yan Liu.
\newblock Invertible image rescaling.
\newblock In {\em {ECCV}}, pages 126--144, 2020.

\bibitem{DBLP:journals/inffus/Xu021}
Han Xu and Jiayi Ma.
\newblock Emfusion: An unsupervised enhanced medical image fusion network.
\newblock {\em Inf. Fusion}, 76:177--186, 2021.

\bibitem{9151265}
Han Xu, Jiayi Ma, Junjun Jiang, Xiaojie Guo, and Haibin Ling.
\newblock U2fusion: {A} unified unsupervised image fusion network.
\newblock {\em {IEEE} TPAMI}, 44(1):502--518, 2022.

\bibitem{xu2020aaai}
Han Xu, Jiayi Ma, Zhuliang Le, Junjun Jiang, and Xiaojie Guo.
\newblock Fusiondn: A unified densely connected network for image fusion.
\newblock In {\em {AAAI}}, pages 12484--12491, 2020.

\bibitem{DBLP:conf/cvpr/Xu0YLL22}
Han Xu, Jiayi Ma, Jiteng Yuan, Zhuliang Le, and Wei Liu.
\newblock Rfnet: Unsupervised network for mutually reinforcing multi-modal
  image registration and fusion.
\newblock In {\em {CVPR}}, pages 19647--19656, 2022.

\bibitem{xu2021drf}
Han Xu, Xinya Wang, and Jiayi Ma.
\newblock Drf: Disentangled representation for visible and infrared image
  fusion.
\newblock {\em IEEE TIM}, 70:1--13, 2021.

\bibitem{xu2021classification}
Han Xu, Hao Zhang, and Jiayi Ma.
\newblock Classification saliency-based rule for visible and infrared image
  fusion.
\newblock {\em IEEE TCI}, 7:824--836, 2021.

\bibitem{xu2021stereo}
Ruikang Xu, Zeyu Xiao, Mingde Yao, Yueyi Zhang, and Zhiwei Xiong.
\newblock Stereo video super-resolution via exploiting view-temporal
  correlations.
\newblock In {\em {ACM} Multimedia}, pages 460--468, 2021.

\bibitem{DBLP:conf/cvpr/Xu0ZSL021}
Shuang Xu, Jiangshe Zhang, Zixiang Zhao, Kai Sun, Junmin Liu, and Chunxia
  Zhang.
\newblock Deep gradient projection networks for pan-sharpening.
\newblock In {\em {CVPR}}, pages 1366--1375, 2021.

\bibitem{DBLP:journals/corr/abs-2005-08448}
Shuang Xu, Zixiang Zhao, Yicheng Wang, Chunxia Zhang, Junmin Liu, and Jiangshe
  Zhang.
\newblock Deep convolutional sparse coding networks for image fusion.
\newblock {\em CoRR}, abs/2005.08448, 2020.

\bibitem{yang2022sir}
Zizheng Yang, Mingde Yao, Jie Huang, Man Zhou, and Feng Zhao.
\newblock Sir-former: Stereo image restoration using transformer.
\newblock In {\em {ACM} Multimedia}, pages 6377--6385, 2022.

\bibitem{yao2019spectral}
Mingde Yao, Zhiwei Xiong, Lizhi Wang, Dong Liu, and Xuejin Chen.
\newblock Spectral-depth imaging with deep learning based reconstruction.
\newblock {\em Optics express}, 27(26):38312--38325, 2019.

\bibitem{DBLP:conf/cvpr/ZamirA0HK022}
Syed~Waqas Zamir, Aditya Arora, Salman Khan, Munawar Hayat, Fahad~Shahbaz Khan,
  and Ming{-}Hsuan Yang.
\newblock Restormer: Efficient transformer for high-resolution image
  restoration.
\newblock In {\em {CVPR}}, pages 5718--5729, 2022.

\bibitem{DBLP:journals/corr/abs-2111-09881}
Syed~Waqas Zamir, Aditya Arora, Salman~H. Khan, Munawar Hayat, Fahad~Shahbaz
  Khan, and Ming{-}Hsuan Yang.
\newblock Restormer: Efficient transformer for high-resolution image
  restoration.
\newblock {\em CoRR}, abs/2111.09881, 2021.

\bibitem{DBLP:journals/ijcv/ZhangM21}
Hao Zhang and Jiayi Ma.
\newblock Sdnet: {A} versatile squeeze-and-decomposition network for real-time
  image fusion.
\newblock {\em Int. J. Comput. Vis.}, 129(10):2761--2785, 2021.

\bibitem{DBLP:conf/aaai/ZhangXXGM20}
Hao Zhang, Han Xu, Yang Xiao, Xiaojie Guo, and Jiayi Ma.
\newblock Rethinking the image fusion: {A} fast unified image fusion network
  based on proportional maintenance of gradient and intensity.
\newblock In {\em {AAAI}}, pages 12797--12804, 2020.

\bibitem{zhang2021deep}
Xingchen Zhang.
\newblock Deep learning-based multi-focus image fusion: A survey and a
  comparative study.
\newblock {\em IEEE TPAMI}, 2021.

\bibitem{DBLP:journals/inffus/ZhangLSYZZ20}
Yu Zhang, Yu Liu, Peng Sun, Han Yan, Xiaolin Zhao, and Li Zhang.
\newblock {IFCNN:} {A} general image fusion framework based on convolutional
  neural network.
\newblock {\em Inf. Fusion}, 54:99--118, 2020.

\bibitem{DBLP:conf/mm/ZhaoZL21}
Jiawei Zhao, Yifan Zhao, and Jia Li.
\newblock {M3TR:} multi-modal multi-label recognition with transformer.
\newblock In {\em {ACM} Multimedia}, pages 469--477, 2021.

\bibitem{DBLP:journals/sigpro/ZhaoXZLZ20}
Zixiang Zhao, Shuang Xu, Chunxia Zhang, Junmin Liu, and Jiangshe Zhang.
\newblock Bayesian fusion for infrared and visible images.
\newblock {\em Signal Process.}, 177:107734, 2020.

\bibitem{zhaoijcai2020}
Zixiang Zhao, Shuang Xu, Chunxia Zhang, Junmin Liu, Jiangshe Zhang, and Pengfei
  Li.
\newblock {DIDFuse}: Deep image decomposition for infrared and visible image
  fusion.
\newblock In {\em {IJCAI}}, pages 970--976, 2020.

\bibitem{DBLP:journals/tcsv/ZhaoXZLZL22}
Zixiang Zhao, Shuang Xu, Jiangshe Zhang, Chengyang Liang, Chunxia Zhang, and
  Junmin Liu.
\newblock Efficient and model-based infrared and visible image fusion via
  algorithm unrolling.
\newblock {\em {IEEE} TCSVT}, 32(3):1186--1196, 2022.

\bibitem{DBLP:journals/corr/abs-2104-06977}
Zixiang Zhao, Jiangshe Zhang, Shuang Xu, Zudi Lin, and Hanspeter Pfister.
\newblock Discrete cosine transform network for guided depth map
  super-resolution.
\newblock In {\em {CVPR}}, pages 5697--5707, June 2022.

\bibitem{DBLP:conf/icmcs/00010XSHL021}
Zixiang Zhao, Jiangshe Zhang, Shuang Xu, Kai Sun, Lu Huang, Junmin Liu, and
  Chunxia Zhang.
\newblock {FGF-GAN:} {A} lightweight generative adversarial network for
  pansharpening via fast guided filter.
\newblock In {\em {ICME}}, pages 1--6, 2021.

\bibitem{DBLP:conf/cvpr/ZhengLZZLWFFXT021}
Sixiao Zheng, Jiachen Lu, Hengshuang Zhao, Xiatian Zhu, Zekun Luo, Yabiao Wang,
  Yanwei Fu, Jianfeng Feng, Tao Xiang, Philip H.~S. Torr, and Li Zhang.
\newblock Rethinking semantic segmentation from a sequence-to-sequence
  perspective with transformers.
\newblock In {\em {CVPR}}, pages 6881--6890, 2021.

\bibitem{DBLP:journals/tgrs/ZhouFH0LW22}
Man Zhou, Xueyang Fu, Jie Huang, Feng Zhao, Aiping Liu, and Rujing Wang.
\newblock Effective pan-sharpening with transformer and invertible neural
  network.
\newblock {\em {IEEE} TGRS}, 60:1--15, 2022.

\bibitem{DBLP:conf/cvpr/ZhouYHYF022}
Man Zhou, Keyu Yan, Jie Huang, Zihe Yang, Xueyang Fu, and Feng Zhao.
\newblock Mutual information-driven pan-sharpening.
\newblock In {\em {CVPR}}, pages 1788--1798, 2022.

\bibitem{DBLP:conf/aaai/ZhuLZLLX19}
Xiaobin Zhu, Zhuangzi Li, Xiaoyu Zhang, Changsheng Li, Yaqi Liu, and Ziyu Xue.
\newblock Residual invertible spatio-temporal network for video
  super-resolution.
\newblock In {\em {AAAI}}, pages 5981--5988, 2019.

\bibitem{DBLP:conf/iclr/ZhuSLLWD21}
Xizhou Zhu, Weijie Su, Lewei Lu, Bin Li, Xiaogang Wang, and Jifeng Dai.
\newblock Deformable {DETR:} deformable transformers for end-to-end object
  detection.
\newblock In {\em {ICLR}}, 2021.

\end{thebibliography}
}

\end{document}